\documentclass[10pt,twocolumn,letterpaper]{article}

\usepackage[pagenumbers]{cvpr} 

\usepackage{amsmath}
\usepackage{amssymb}
\usepackage{booktabs}
\usepackage{color}
\usepackage{colortbl}
\usepackage{graphicx}
\usepackage{enumitem}
\usepackage{tablefootnote}
\usepackage{makecell}
\usepackage{caption}
\usepackage{multirow}
\usepackage{xcolor}
\usepackage{subcaption}
\usepackage[ruled]{algorithm2e}
\usepackage{float}
\usepackage[symbol]{footmisc}

\SetCommentSty{mycommfont}
\definecolor{cyan}{cmyk}{.3,0,0,0}
\definecolor{lightblue}{HTML}{B0C4DE}
\definecolor{darkblue}{HTML}{177CB0}
\definecolor{ggreen}{HTML}{228B22}
\definecolor{gorange}{HTML}{D84F14}
\definecolor{gblue}{HTML}{0072BD}
\newcommand{\red}[1]{\textcolor{red}{#1}}

\usepackage{amsmath,amsfonts,bm}

\def\eqref#1{equation~\ref{#1}}

\def\1{\bm{1}}

\def\eps{{\epsilon}}

\def\vf{{\bm{f}}}
\def\vg{{\bm{g}}}
\def\vh{{\bm{h}}}

\def\vp{{\bm{p}}}
\def\vq{{\bm{q}}}

\def\vu{{\bm{u}}}
\def\vv{{\bm{v}}}

\def\mP{{\bm{P}}}
\def\mQ{{\bm{Q}}}

\def\mU{{\bm{U}}}
\def\mV{{\bm{V}}}

\DeclareMathAlphabet{\mathsfit}{\encodingdefault}{\sfdefault}{m}{sl}
\SetMathAlphabet{\mathsfit}{bold}{\encodingdefault}{\sfdefault}{bx}{n}

\newcommand{\R}{\mathbb{R}}

\usepackage[pagebackref,breaklinks,colorlinks]{hyperref}

\def\ourablmethod{{nCLIP}\xspace} 
\def\oursecablmethod{{yCLIP}\xspace} 
\def\ourmethod{{xCLIP}\xspace} 

\usepackage[capitalize]{cleveref}
\crefname{section}{Sec.}{Secs.}
\Crefname{section}{Section}{Sections}
\Crefname{table}{Table}{Tables}
\crefname{table}{Tab.}{Tabs.}

\usepackage{xspace}
\makeatletter
\DeclareRobustCommand\onedot{\futurelet\@let@token\@onedot}
\def\@onedot{\ifx\@let@token.\else.\null\fi\xspace}

\def\eg{\emph{e.g}\onedot} 
\def\ie{\emph{i.e}\onedot} 
 
\def\etc{\emph{etc}\onedot}

\makeatother

\def\cvprPaperID{*****} 
\def\confName{CVPR}
\def\confYear{2022}

\begin{document}

\title{Non-Contrastive Learning Meets Language-Image Pre-Training}

\author{
	Jinghao Zhou \quad Li Dong \quad Zhe Gan \quad Lijuan Wang \quad Furu Wei \\
	Microsoft
}

\maketitle

\begin{abstract}
Contrastive language-image pre-training (CLIP) serves as a de-facto standard to align images and texts. Nonetheless, the loose correlation between images and texts of web-crawled data renders the contrastive objective data inefficient and craving for a large training batch size.
In this work, we explore the validity of non-contrastive language-image pre-training (\ourablmethod), and study whether nice properties exhibited in visual self-supervised models can emerge. We empirically observe that the non-contrastive objective nourishes representation learning while sufficiently underperforming under zero-shot recognition.
Based on the above study, we further introduce \ourmethod, a multi-tasking framework combining CLIP and \ourablmethod, and show that \ourablmethod aids CLIP in enhancing feature semantics. The synergy between two objectives lets \ourmethod{} enjoy the best of both worlds: superior performance in both zero-shot transfer and representation learning.
Systematic evaluation is conducted spanning a wide variety of downstream tasks including zero-shot classification, out-of-domain classification, retrieval, visual representation learning, and textual representation learning, showcasing a consistent performance gain and validating the effectiveness of \ourmethod. 
\end{abstract}


\section{Introduction}
\label{sec:introduction}

Language-image pre-training which simultaneously learns textual and visual representation from large-scale image-text pairs has revolutionized the field of representation learning~\cite{virtex,clip}, vision-language understanding~\cite{uniter}, and text-to-image generation~\cite{dalle}. Compared to traditional visual models, language-instilled ones intrinsically inherit the capability of zero-shot or few-shot learning prominently demonstrated by large language models such as GPT-3~\cite{gpt3}. The precursor system, \textbf{C}ontrastive \textbf{L}anguage-\textbf{I}mage \textbf{P}re-Training~\cite{clip} (\textbf{CLIP}) that explicitly aligns the projected features of two modalities, has demonstrated surprising capabilities of zero-shot, representation learning, and robustness, being applied to a wide range of fields~\cite{vild,dalle2,clipasso,dbot}.

To learn from noisy web-crawled image-text pairs, CLIP adopts a formulation of the contrastive objective, where the image and text within a pair are considered as unique instances and are encouraged to be discriminated from all the other negative instances. However, web-crawled image-text pairs~\cite{cc3m,yfcc15m} are usually loosely correlated in the sense that one caption (image) can match reasonably with multiple images (captions) besides the ground-truth one, as statistically shown in~\cite{otter}. Hence, it is inaccurate and data-inefficient for representation learning to neglect other sensible matches and overlook the semantics hidden inside the textual description. This is also solidified by the undesirable discriminative ability or transferring performance of the visual encoder pre-trained under the contrastive objective, as suggested in~\cite{unicl}. Mining semantics from plentiful concepts appearing in captions, therefore, solicits further exploration beyond the vanilla contrastive objective.

\begin{figure}[!t]
\centering
\includegraphics[height=0.49\linewidth]{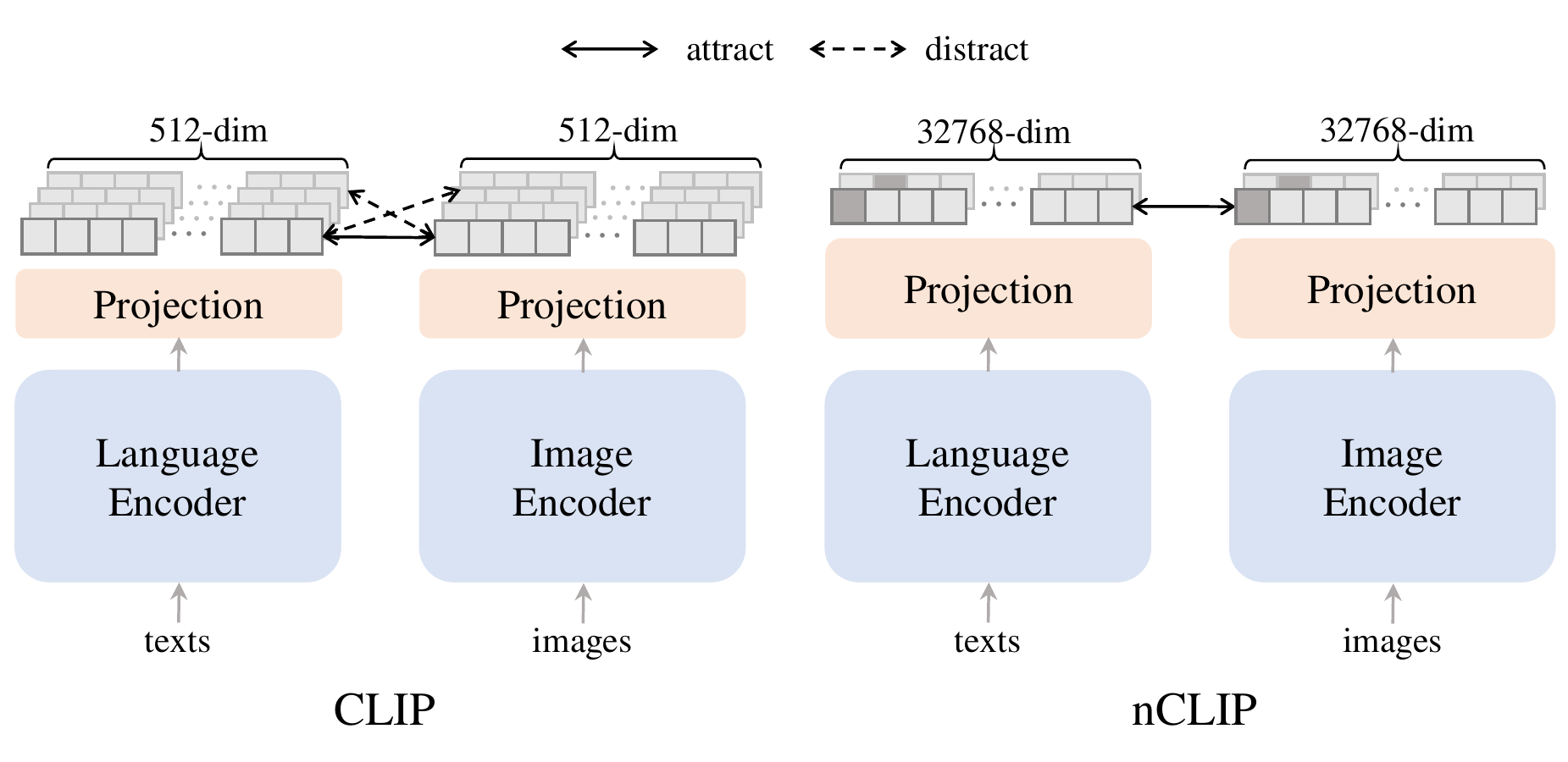}
\vspace{-0.7cm}
\caption{\textbf{Architecture comparison between CLIP and \ourablmethod.}  We take \textbf{base}-size encoders for an instance. CLIP discriminates instances within a batch using 512-dim projected embeddings. \ourablmethod projects each modality into a 32768-dim probability distribution as the pseudo-label to supervise the prediction from the other modality. Darker blocks of \ourablmethod depict a higher response for cluster distribution, signifying the clusters to which text or image instances may belong. \ourmethod is a multi-tasking framework with both CLIP and \ourablmethod.}
\label{fig:arch}
\end{figure}

Previously, some works resort to mining nearest neighbor positive samples via intra-modality similarity~\cite{otter,declip} but require extra storage for auxiliary modules, \ie, the teacher network~\cite{otter} or the memory bank~\cite{declip}. Other works conduct multi-tasking with image-label supervision~\cite{unicl,basic,coca}, transforming the captions into tag format or the image tag into the captioning format for unified contrastive learning. Notwithstanding, such practice is still unable to account for the loose assignment within the image-text captioning pairs. 
Intuitively, a caption for an image usually identifies existing objects within the image, which can be captured by \textit{a probabilistic distribution estimating how the text is assigned to one or multiple object clusters}~\cite{swav,deepcluster}. Such estimation depicts the semantic meanings for its contained visual contents, and thus can be leveraged as a pseudo-label to guide the learning of the visual encoder. 
 
Inspired by the superiority of visual self-supervised learning (SSL) models pre-trained with the non-contrastive objective~\cite{dino,swav},  we explore whether the non-contrastive objective can be used across modalities for pre-training language-image models, and whether nice properties displayed on visual SSL models can be inherited. 
To this end, we follow the same setup as CLIP except for the objective, and study \textbf{n}on-\textbf{C}ontrastive \textbf{L}anguage-\textbf{I}mage \textbf{P}re-Training (\textbf{\ourablmethod}). For an image-text pair, we use the estimation of the textual (visual) distribution as the target to supervise the visual (textual) distribution, measured by cross-entropy. Additional regularizers are applied to avoid trivial solutions. Schematic comparison between CLIP and \ourablmethod is shown in~\cref{fig:arch}.
Theoretically, such formulation takes one modality and the cluster centroid that the other modality belongs to as the positive samples to contrast~\cite{swav}, as opposed to direct features of two modalities as in contrastive learning, such that a single image is tolerated to be aligned with multiple captions and concepts.
Based on the above formulation, we conduct a systematic study in terms of zero-shot transfer and representation learning. We empirically observe that, while nCLIP demonstrates desirable representation learning capability for both modalities, it underperforms prominently in zero-shot classification and retrieval tasks where models pre-trained with negative samples naturally prevail.

Seeking for unique strengths of both worlds, we further perform multi-tasking of CLIP and \ourablmethod, short as \textbf{\ourmethod}, and seek synergy between two distinct pre-training objectives.
With the bearable add-on of computational resources (\eg, $\sim27\%$ memory-wise and $\sim30\%$ time-wise), \ourmethod{} achieves consistent performance gain compared to CLIP on a wide range of tasks spanning from zero-shot classification, retrieval, linear probing, and fine-tuning, \etc. An extensive study with different pre-training datasets, evaluation metrics, and optimization configurations is conducted to validate \ourmethod's effectiveness in mining semantics and improving data efficiency. Particularly, the base-size model pre-trained using \ourmethod{} under 35-million publicly available image-text pairs achieves a performance gain of \textbf{3.3}\% and \textbf{1.5}\% on an average of 27 classification tasks~\cite{clip}, in terms of zero-shot classification and linear probing accuracy, respectively. Performance gain is also valid in terms of zero-shot retrieval, semi-supervised learning, and fine-tuning, with \textbf{3.7} points of R@1 on Flickr30K~\cite{flickr30k}, \textbf{1.7}\% accuracy on 1\% subset of ImageNet~\cite{imagenet}, and \textbf{0.3}\% accuracy on ImageNet, respectively.

\section{Related Work}
\label{sec:relatedwork}

\paragraph{Language-image pre-training.} Language-image pre-training, \textit{aka} vision-language pre-training (VLP), learns to jointly cope with vision and language input using multi-modality models. Early practices~\cite{lxmert,visualbert,vlbert,uniter} separately encode modalities, \ie, texts with linear embedding and images with convolutional neural networks and region proposals~\cite{fasterrcnn}. Recent works take direct images instead of pre-fetched region features as input, using \eg, dual-encoder architecture for alignment~\cite{align,clip,basic}, single-encoder architecture for fusion~\cite{vilt,simvlm}, or their combinatorial practices forming an encoder-decoder architecture~\cite{albef,coca}. Most recent literature aims to expand as many supported transferable downstream tasks as possible via pre-training large-scale foundation models~\cite{flamingo,coca}. 
Out of previous works, CLIP~\cite{clip} blazes the trail to learn transferable visual features from text supervision, and demonstrates surprising zero-shot recognition results. Follow-up study includes leveraging dense~\cite{filip}, augmentation~\cite{uniclip}, and uni-modal self-supervised~\cite{declip,slip,otter} signals for improved performance. In this work, we base our study on the dual-encoder architecture, and pre-train models either with the \textit{contrastive} objective as CLIP~\cite{clip}, or the \textit{non-contrastive} objective. Further, we seek their complementarity and study whether synergy between two objectives exists. 
 
\paragraph{Contrastive learning.} The idea of contrastive learning is popularized under visual self-supervised learning (SSL) which aims at learning transferable visual representation from unlabelled images. In common practices~\cite{instdist,simclr,moco}, each image is considered a single class, attracted to its augmented views while pulled away from views of other images. The contrastive objective is proved robust and effective in a wide range of realms beyond visual understanding, including representation learning for natural language~\cite{simcse}, audio~\cite{cdpam}, structured input~\cite{c-swm}, robotics~\cite{curl}, and multi-modality~\cite{clip,ccl}. However, a caveat of these approaches is the requirement for large batch size~\cite{simclr} or memory bank~\cite{instdist,moco}, which lies intrinsic in the formulation of InfoNCE estimator~\cite{infomax} and plagues the pre-training of large-size models due to hardware limitations.  

\paragraph{Non-contrastive learning.} Beyond the contrastive paradigm, recent state of the arts from visual SSL manage to relieve the dependency on negative samples. With different subtleties to avert collapsing solutions, these works can optimize the affinity of augmented representations alone and are categorized as \textit{non-contrastive} framework. To avoid model collapsing, common practices include asymmetrical architecture~\cite{byol,simsiam}, dimension de-correlation~\cite{barlowtwins,wmse,vicreg}, and clustering~\cite{deepcluster,swav,dino,selfclassifier,msn,twist}. 
Our approach takes inspiration from the last category since it intrinsically frees the absoluteness of assignment of positive samples. Moreover, rich semantics induced by explicit clustering can automatically group visual concepts from noisy image-text pairs, facilitating models' representation learning. 

\section{Approach}
\label{sec:approach}

\subsection{Framework}
\label{sec:framework}

Suppose $\vg$ and $\vh \in \mathbb{R}^{D\times1}$ are projected backbone features of two modalities. $D$ is the feature dimension. We start by formulating the objective of dominant \textit{contrastive} framework CLIP, followed by the introduced \textit{non-contrastive} framework \ourablmethod. The final framework \ourmethod{} takes the multi-tasking of two objectives, the complete computational pipeline of which is shown in \cref{alg:xclip}.
 
\paragraph{Contrastive pre-training~\cite{clip}.}
Let $\vu = \vg / ||\vg||$ and $\vv = \vh / ||\vh||$. $\mU = [\vu_1, \vu_2, ..., \vu_B] \in \mathbb{R}^{D\times B}$, $\mV = [\vv_1, \vv_2, ..., \vv_B] \in \mathbb{R}^{D\times B}$ are concatenated embeddings over the batch. $B$ is the batch size. The contrastive objective is formulated as
\begin{align}
\label{eq:cliploss}
\mathcal{L}_\mathrm{CLIP} &= \mathrm{InfoNCE}(\mU^\mathrm{T} \mV / \sigma) \nonumber \\[5pt]
& = - \frac{1}{B} \sum_{i\in B} \ \mathrm{log}\frac{\mathrm{exp}(\vu_i^\mathrm{T}\vv_i/ \sigma)}{\sum_{j\in B}\mathrm{exp}(\vu_i^\mathrm{T}\vv_j/ \sigma)},
\end{align}
where $\sigma$ is a trainable parameter controlling the temperature. $\mathcal{L}_\mathrm{CLIP}$ is symmetrized by setting $\vg$ and $\vh$ as projected features of the images and texts by turns and taking the average of two terms. Vanilla $\mathrm{InfoNCE}$ can be decoupled into two terms accounting for affinity and variability separately~\cite{barlowtwins}. While both variability terms in the contrastive and non-contrastive objectives require for batch statistics (\eg, $\mathcal{L}_\mathrm{EH}$ in~\cref{eq:ncliploss}), CLIP formulation explicitly maximizes the distance between negative pairs of batch samples. Hence, it degrades the learning performance and data efficiency when models are pre-trained with noisy data where sensible matches occur within negative pairs.

\paragraph{Non-contrastive pre-training.}
Let $\vp=\mathrm{softmax}(\vg)$ and $\vq=\mathrm{softmax}(\vh)$. We transform the projected feature into probability distributions, which can be seen as an assignment over semantic clusters (\ie, projection weights). We take $\vp$ as the target distribution and learn to estimate the predicted distribution $\vq$ by minimizing their cross-entropy:
\begin{align}
\label{eq:celoss}
\mathcal{L}_\mathrm{CE} &= -\vp^\mathrm{T}\mathrm{log}(\vq).
\end{align}
Note the target and prediction branch are backpropagated at the same time and $\mathcal{L}_\mathrm{CE}$ is symmetrized as $\mathcal{L}_\mathrm{CLIP}$. To avert collapsing solutions, we incorporate entropy regularizers, \ie, entropy minimization~\cite{twist} and mean entropy maximization~\cite{paws,msn}, via minimizing:
\begin{align}
\label{eq:ehheloss}
\mathcal{L}_\mathrm{EH} = -\vp^\mathrm{T}\mathrm{log}(\vp), \ -\mathcal{L}_\mathrm{HE} = \overline\vp^\mathrm{T}\mathrm{log}(\overline\vp),
\end{align}
where $\overline\vp = \mathbb{E}(p) \approx \frac{1}{B}\sum^B_{i=1}\vp_i$ is the average distribution over batch. $\mathcal{L}_\mathrm{EH}$ and $\mathcal{L}_\mathrm{HE}$ are symmetrized by taking the average with another term where $\vp$ is replaced by $\vq$. $\mathcal{L}_\mathrm{EH}$ encourages the model to make deterministic predictions and ensures assignment's \textit{sharpness}. $\mathcal{L}_\mathrm{HE}$ encourages the model to utilize a full set of projection weights and ensures the assignment's \textit{smoothness}. We illustrate in \cref{sec:ablation} and \cref{sec:methodcompare} that $\mathcal{L}_\mathrm{EH}$ and $\mathcal{L}_\mathrm{HE}$ are minimally sufficient to yield non-collapsing solutions. We empirically show in \cref{tab:methodcompare} that the solutions is non-trivial as long as both \textit{sharpness} and \textit{smoothness} are guaranteed, \eg, via Sinkhorn algorithm~\cite{swav} instead of loss regularizers. We opt for \cref{eq:ehheloss} for its simplicity and consistency between pre-training and evaluation. The overall non-contrastive objective is formulated as
\begin{align}
\label{eq:ncliploss}
\mathcal{L}_\mathrm{\ourablmethod} &= \mathcal{L}_\mathrm{CE} + \lambda_1 \cdot \mathcal{L}_\mathrm{EH} - \lambda_2 \cdot \mathcal{L}_\mathrm{HE},
\end{align}
where $\lambda_1$ and $\lambda_2$ controls the weight of the regularization. We empirically find that setting $\lambda_1=\lambda_2-1=0.5$ yields stable training and favorable transfer performances. We detail our findings in \cref{sec:ablation} and \cref{sec:entropyreg}.
During evaluation, we utilize negative cross-entropy $-\mathcal{L}_\mathrm{CE}$ as the similarity metric for zero-shot transfer experiments.
As schematically showcased in \cref{fig:arch}, \ourablmethod does not reply on negative examples but requires a relatively larger projection head and greater output dimension. 
Comparing \ourablmethod with CLIP, we empirically find that \ourablmethod produces more coarse-grained (\eg, zero-shot retrieval in \cref{sec:zeroshot}) but semantic-rich (\eg, linear probing in~\cref{sec:representation}) projections. Empirical evidence suggests that the model will fail to deliver reasonable zero-shot and retrieval results if pre-trained without negative pairs (\eg, \ourablmethod's 25.2 \textit{vs.} CLIP's 73.8 points of R@1 under Flickr30K I$\rightarrow$T retrieval as shown in~\cref{tab:retrieval}). 

\setlength{\textfloatsep}{5pt}
\begin{algorithm}[!tbp]
\begin{minipage}{1.0\textwidth}
\BlankLine
\DontPrintSemicolon
\SetKwProg{Fn}{def}{:}{}

\tcp{enc, proj: encoder \& projector}

$\vf_I$, $\vf_T =$ enc(img), enc(text) 

$\vg_I$, $\vg_T =$ proj$_\mathrm{CLIP}(\vf_I)$, proj$_\mathrm{CLIP}(\vf_T)$

$\vh_I$, $\vh_T =$ proj$_\mathrm{\ourablmethod}(\vf_I)$, proj$_\mathrm{\ourablmethod}(\vf_T)$

$\mathcal{L}_\mathrm{\ourmethod} = \lambda_\mathrm{CLIP} \cdot \mathcal{L}_\mathrm{CLIP}(\vg_I, \vg_T, \sigma)$

\hskip3.5em $+ \lambda_\mathrm{\ourablmethod} \cdot \mathcal{L}_\mathrm{\ourablmethod}(\vh_I, \vh_T, \lambda_1, \lambda_2)$

\KwRet $\mathcal{L}_\mathrm{\ourmethod}$
\BlankLine

\SetKwFunction{FMain}{$\mathcal{L}_\mathrm{CLIP}$}
\Fn{\FMain{$\vf_I$, $\vf_T$, $\sigma$}}{
    
    \tcp{INCE: InfoNCE with \cref{eq:cliploss}}
    
    
    $\vu_I$ = normalize($\vf_I$, p=2, dim=1)
    
    $\vu_T$ = normalize($\vf_T$, p=2, dim=1)
    
    $\mathcal{L}$ = $\mathrm{INCE}(\mU_I^\mathrm{T}\mU_T / \sigma$) + $\mathrm{INCE}(\mU_T^\mathrm{T}\mU_I / \sigma$)
    
    \KwRet $\mathcal{L}$ / 2
    \BlankLine
}
\BlankLine
\SetKwFunction{FMain}{$\mathcal{L}_\mathrm{\ourablmethod}$}
\Fn{\FMain{$\vf_I$, $\vf_T$, $\lambda_1$, $\lambda_2$}}{

    
    $\vp_I$ = softmax($\vf_I$, dim=1)
    
    $\vp_T$ = softmax($\vf_T$, dim=1)
    
    $\mathcal{L}_{\mathrm{CE}}$ = - ($\vp_I \cdot$ log($\vp_T$) + $\vp_T \cdot$ log($\vp_I$)).sum(dim=1)
    
    \hskip3em .mean(dim=0)

    $\mathcal{L}_{\mathrm{EH}}$ = - ($\vp_I \cdot$ log($\vp_I$) + $\vp_T \cdot$ log($\vp_T$)).sum(dim=1)
    
    \hskip3em .mean(dim=0)

    $\overline \vp_I$, $\overline \vp_T$ = $\vp_I$.mean(dim=0), $\vp_T$.mean(dim=0)

    $\mathcal{L}_{\mathrm{HE}}$ = - ($\overline \vp_I \cdot$ log($\overline \vp_I$) + $\overline \vp_T \cdot$ log($\overline \vp_T$)).sum(dim=1)

    $\mathcal{L}$ = $\mathcal{L}_{\mathrm{CE}}$ + $\lambda_1 \cdot \mathcal{L}_{\mathrm{EH}}$ - $\lambda_2 \cdot \mathcal{L}_{\mathrm{HE}}$
    
    \KwRet $\mathcal{L}$ / 2
    \BlankLine
}
\caption{PyTorch-like pseudo-code of \ourmethod.}
\label{alg:xclip}
\end{minipage}
\end{algorithm}

\paragraph{Contrastive meets non-contrastive.} Given that two objectives each have their own congenital defects, we further seek the complementarity between $\mathcal{L}_\mathrm{CLIP}$ and $\mathcal{L}_\mathrm{\ourablmethod}$, and pre-train the models with both objectives, written as
\begin{align}
\label{eq:xcliploss}
\mathcal{L}_\mathrm{\ourmethod} &= \lambda_\mathrm{CLIP} \cdot \mathcal{L}_\mathrm{CLIP} + \lambda_\mathrm{\ourablmethod} \cdot \mathcal{L}_\mathrm{\ourablmethod},
\end{align}
where $\lambda_\mathrm{CLIP}$ and $\lambda_\mathrm{\ourablmethod}$ control the weight of the objectives. During evaluation, we find using the negative cosine metric with CLIP's projection head for zero-shot transfer experiments generally leads to better results.  
We note that it is not immediately clear that the two objectives certainly induce stronger models, since they build qualitatively distinctive latent spaces. Qualitatively, we show in~\cref{sec:experiments} that \ourablmethod helps CLIP to encode semantics which intrinsically lacks in CLIP, while CLIP helps \ourablmethod to be transferable for zero-shot recognition. The synergy between the two objectives prompts consistent performance gains across a wide range of tasks by \ourmethod{} over CLIP.

\subsection{Implementation}
\label{sec:implementation}

\begin{table*}[!tbp]
\centering
\footnotesize
\setlength{\tabcolsep}{0.52mm}{
\begin{tabular}{cccccccccccccccccccccccccccc|c}
Model & \rotatebox{90}{\scriptsize Food101} &  \rotatebox{90}{\scriptsize CIFAR10} & \rotatebox{90}{\scriptsize CIFAR100} & \rotatebox{90}{\scriptsize Birdsnap} & \rotatebox{90}{\scriptsize SUN397} & \rotatebox{90}{\scriptsize Cars} & \rotatebox{90}{\scriptsize Aircraft} & \rotatebox{90}{\scriptsize VOC2007} & \rotatebox{90}{\scriptsize DTD} & \rotatebox{90}{\scriptsize Pets} & \rotatebox{90}{\scriptsize Caltech101} & \rotatebox{90}{\scriptsize Flowers} & \rotatebox{90}{\scriptsize MNIST} & \rotatebox{90}{\scriptsize FER2013} & \rotatebox{90}{\scriptsize STL10} & \rotatebox{90}{\scriptsize EuroSAT} & \rotatebox{90}{\scriptsize RESISC45} & \rotatebox{90}{\scriptsize GTSRB} & \rotatebox{90}{\scriptsize KITTI} & \rotatebox{90}{\scriptsize Country211} & \rotatebox{90}{\scriptsize PCAM} & \rotatebox{90}{\scriptsize UCF101} & \rotatebox{90}{\scriptsize Kinetics700} & \rotatebox{90}{\scriptsize CLEVR} & \rotatebox{90}{\tiny HatefulMemes} &  \rotatebox{90}{\scriptsize SST} & \rotatebox{90}{\scriptsize \bf ImageNet} & \rotatebox{90}{\scriptsize \bf Average} \\
\toprule
CLIP & 61.2 & 79.7 & 50.6 & 23.7 & 56.5 & 15.9 & 5.8 & 46.4 & 27.6 & 54.7 & 71.3 & 48.9 & 10.5 & 37.3 & 91.3 & 24.5 & 39.7 & 13.1 & 31.6 & 9.1 & 50.0 & 45.0 & 32.4 & 12.8 & \textbf{53.0} & 49.1 & 45.7 & 40.3 \\
\ourablmethod & 28.4 & 79.5 & 49.1 & 11.3 & 57.0 & 5.9 & 4.5 & 51.4 & 22.9 & 14.6 & 65.0 & 23.1 & 9.9 & 13.5 & \textbf{94.8} & 15.1 & 21.2 & 2.7 & 35.4 & 5.8 & \textbf{51.2} & 42.0 & 28.4 & 12.4 & 52.7 & 50.0 & 37.0 & 32.7 \\
\bf \ourmethod{} & \textbf{65.8} & \textbf{83.4} & \textbf{54.5} & \textbf{25.1} & \textbf{59.9} & \textbf{18.0} & \textbf{5.8} & \textbf{52.2} & \textbf{33.2} & \textbf{57.1} & \textbf{73.9} & \textbf{50.0} & \textbf{12.3} & \textbf{39.0} & 92.8 & \textbf{40.0} & \textbf{43.6} & \textbf{16.3} & \textbf{39.8} & \textbf{9.3} & 51.1 & \textbf{49.8} & \textbf{35.4} & \textbf{18.4} & 52.5 & \textbf{50.2} & \textbf{48.8} & \textbf{43.6} \\
\bottomrule
\end{tabular}}
\vspace{-0.3cm}
\caption{\textbf{Zero-shot classification.} We report on a variety of classification benchmarks with ViT-B/16 pre-trained on IT35M. Detailed protocols for each dataset strictly follow CLIP~\cite{clip}. \ourmethod{} achieves a consistent performance gain compared to CLIP in a wide range of classification datasets. Best results of each column are \textbf{bolded}.}
\label{tab:zeroshot}
\end{table*}

\paragraph{Architecture.} We train the base-size model with ViT-B/16~\cite{vit} as the visual encoder. The model configuration of the base-size text encoder follows CLIP~\cite{clip} with byte-pair encoding (BPE) and a maximum context length of 77. The projection head generating output for the non-contrastive objective is a 2-layer MLP with 4096-dim hidden layers, GELU~\cite{gelu}, and BatchNorm~\cite{batchnorm}. The last layer is of 32768 output dimension and followed by a BatchNorm without affine transformation. The projection head for the contrastive objective is a single linear layer with no bias and a dimension of 512.

\paragraph{Pre-training data.} We train our method with publicly available datasets: COCO~\cite{coco}, Visual Genome~\cite{vg}, SBU Captions~\cite{sbu}, Conceptual Caption 3M~\cite{cc3m}, Conceptual Caption 12M~\cite{cc12m}, and filtered 14M-size subset of Yahoo Flickr Creative Commons 100M dataset, consisting a total of 35M \underline{I}mage-\underline{T}ext pairs (IT35M). 
We are also intrigued about the behaviors of models pre-trained on ImagNet-21K~\cite{imagenet} (IN21K) dataset taking label names as annotation texts by concatenating them with a sampled prompt, similar to~\cite{basic,coca}. To notify, ImageNet-21K is a subset of ImageNet-22K with classes of ImageNet-1K excluded for fair downstream evaluation.
We detail the models' data scaling behavior in \cref{tab:datascale}.

\paragraph{Optimization.} We use a batch size of 4096, with all data distributed across 32 V100 GPUs. Models are trained by default with AdamW~\cite{adamw} optimizer, a peak learning rate of $1e^{-3}$, cosine scheduler, a weight decay of 0.2, a $\beta_2$ for AdamW of 0.98, a $\epsilon$ for AdamW of $1e^{-6}$, and automatic mixed-precision for 3 warm-up epochs and a total of 32 epochs. $\lambda_1$ and $\lambda_2$ are set as 0.5 and 1.5, respectively. The two objectives are multi-tasked with $\lambda_\textrm{nCLIP}$ being 1 and $\lambda_\textrm{CLIP}$ being 0.2. Augmentations and pre-processing of pre-training data follow CLIP with image size randomly cropped and resized within the scale of (0.5, 1.0). Detailed pre-training hyper-parameters are detailed in \cref{sec:hyperparam}. 


\section{Experiments}
\label{sec:experiments}

\begin{table}[!tbp]
\centering
\setlength{\tabcolsep}{1.4mm}{
\begin{tabular}{cccccccc|c}
Model & \rotatebox{90}{\footnotesize IN-A} &  \rotatebox{90}{\footnotesize IN-R} & \rotatebox{90}{\footnotesize IN-v2} & \rotatebox{90}{\footnotesize IN-Ske} & \rotatebox{90}{\footnotesize IN-Sty} & \rotatebox{90}{\footnotesize IN-C} & \rotatebox{90}{\footnotesize ObjNet} & \rotatebox{90}{\footnotesize \bf Average} \\
\toprule
CLIP & 21.3 & 43.7 & 37.9 & 26.5 & 4.9 & 11.9 & 15.5 & 23.1 \\
\ourablmethod & 21.5 & 29.0 & 30.8 & 17.7 & 3.6 & 11.5 & 12.0 & 18.0 \\
\bf \ourmethod{} & \textbf{27.7} & \textbf{49.1} & \textbf{40.6} & \textbf{30.6} & \textbf{6.1} & \textbf{14.7} & \textbf{17.8} & \textbf{26.7} \\
\bottomrule
\end{tabular}}
\vspace{-0.2cm}
\caption{\textbf{Zero-shot out-of-distribution classification.} We report on a variety of out-of-distribution classification benchmarks. \ourmethod{} demonstrates stronger robustness on various out-of-domain classification datasets.}
\label{tab:oodzeroshot}
\end{table}

\begin{table}[tbp]
\centering
\setlength{\tabcolsep}{1.3mm}{
\begin{tabular}{ccccccc}
\multirow{2}{*}{Model} & \multicolumn{3}{c}{NUS-WIDE} & \multicolumn{3}{c}{OpenImages} \\
\cmidrule(lr){2-4}\cmidrule(lr){5-7}
& mAP & F1@3 & F1@5 & mAP & F1@10 & F1@20  \\
\toprule
CLIP & 15.1 & 33.3 & 16.0 & 81.1 & 13.4 & 7.2 \\
\ourablmethod & \textbf{16.1} & \textbf{35.6} & 16.6 & \textbf{81.4} & 11.6 & 6.4 \\
\bf \ourmethod{} & 15.3 & 35.2 & \textbf{16.8} & 81.2 & \textbf{13.8} & \textbf{7.4} \\
\bottomrule
\end{tabular}}
\vspace{-0.2cm}
\caption{\textbf{Zero-shot multi-label classification.} We report mAP and F1 scores with ViT-B/16 pre-trained on IT35M. \ourablmethod showcases the best multi-label classification capability. }
\label{tab:multilabel}
\end{table}

\begin{table*}[tbp]
\centering
\setlength{\tabcolsep}{2.6mm}{
\begin{tabular}{ccccccccccccc}
\multirow{3}{*}{Model} & \multicolumn{6}{c}{Flickr30K} & \multicolumn{6}{c}{MSCOCO} \\
\cmidrule(lr){2-7}\cmidrule(lr){8-13}
& \multicolumn{3}{c}{I$\rightarrow$T} & \multicolumn{3}{c}{T$\rightarrow$I} & \multicolumn{3}{c}{I$\rightarrow$T} & \multicolumn{3}{c}{T$\rightarrow$I} \\
\cmidrule(lr){2-4}\cmidrule(lr){5-7}\cmidrule(lr){8-10}\cmidrule(lr){11-13}
& R@1 & R@5 & R@10 & R@1 & R@5 & R@10 & R@1 & R@5 & R@10 & R@1 & R@5 & R@10 \\
\toprule
CLIP & 73.8 & 93.3 & 97.1 & 52.9 & 79.3 & 87.7 & 59.4 & 85.2  & 91.9 & 34.3 & 62.7 & 74.0 \\
\ourablmethod & 26.2 & 53.3 & 67.9 & 20.1 & 48.5 & 61.1 & 21.4 & 47.9 & 61.6 & 13.4 & 37.1 & 50.3 \\
\bf \ourmethod{} & \textbf{77.5} & \textbf{95.6} & \textbf{97.7} & \textbf{57.3} & \textbf{82.9} & \textbf{89.0} & \textbf{63.3} & \textbf{87.5} & \textbf{94.1} & \textbf{38.4} & \textbf{66.0} & \textbf{76.7} \\
\bottomrule
\end{tabular}}
\vspace{-0.2cm}
\caption{\textbf{Zero-shot image-to-text retrieval.} We report R@1, R@5, and R@10 in both image-to-text (I$\rightarrow$T) and text-to-image (T$\rightarrow$I) settings with ViT-B/16 pre-trained on IT35M. \ourmethod{} consistently improves CLIP in retrieval. }
\vspace{-0.3cm}. 
\label{tab:retrieval}
\end{table*}

\begin{table*}[tbp]
\centering
\footnotesize
\setlength{\tabcolsep}{0.45mm}{
\begin{tabular}{cccccccccccccccccccccccccccc|c}
Model & \rotatebox{90}{\scriptsize Food101} &  \rotatebox{90}{\scriptsize CIFAR10} & \rotatebox{90}{\scriptsize CIFAR100} & \rotatebox{90}{\scriptsize Birdsnap} & \rotatebox{90}{\scriptsize SUN397} & \rotatebox{90}{\scriptsize Cars} & \rotatebox{90}{\scriptsize Aircraft} & \rotatebox{90}{\scriptsize VOC2007} & \rotatebox{90}{\scriptsize DTD} & \rotatebox{90}{\scriptsize Pets} & \rotatebox{90}{\scriptsize Caltech101} & \rotatebox{90}{\scriptsize Flowers} & \rotatebox{90}{\scriptsize MNIST} & \rotatebox{90}{\scriptsize FER2013} & \rotatebox{90}{\scriptsize STL10} & \rotatebox{90}{\scriptsize EuroSAT} & \rotatebox{90}{\scriptsize RESISC45} & \rotatebox{90}{\scriptsize GTSRB} & \rotatebox{90}{\scriptsize KITTI} & \rotatebox{90}{\scriptsize Country211} & \rotatebox{90}{\scriptsize PCAM} & \rotatebox{90}{\scriptsize UCF101} & \rotatebox{90}{\scriptsize Kinetics700} & \rotatebox{90}{\scriptsize CLEVR} & \rotatebox{90}{\tiny HatefulMemes} &  \rotatebox{90}{\scriptsize SST} & \rotatebox{90}{\scriptsize \bf ImageNet} & \rotatebox{90}{\scriptsize \bf Average} \\
\toprule
CLIP & 83.3 & 92.0 & 75.6 & 54.4 & 78.8 & 60.5 & 33.8 & 91.1 & 72.4 & 79.0 & 90.0 & 96.1 & 96.2 & 56.5 & 97.1 & 94.8 & 90.3 & 77.3 & \textbf{69.7} & 23.2 & 83.4 & 78.9 & 53.9 & \textbf{50.3} & 55.9 & \textbf{55.8} & 71.4 & 72.7 \\
\ourablmethod & 83.8 & 93.2 & 76.3 & \textbf{66.8} & 79.3 & 49.3 & 28.0 & 90.1 & 70.6 & 70.2 & 90.5 & 95.0 & 95.4 & 54.7 & 97.4 & 91.9 & 88.2 & 75.7 & 64.0 & 22.7 & \textbf{84.2} & 79.9 & 51.0 & 44.6 & 55.6 & 55.2 & 73.9 & 71.4 \\
\bf \ourmethod{} & \textbf{85.3} & \textbf{93.4} & \textbf{77.8} & 58.4 & \textbf{80.7} & \textbf{62.3} & \textbf{36.7} & \textbf{92.3} & \textbf{74.0} & \textbf{81.5} & \textbf{91.6} & \textbf{97.0} & \textbf{96.8} & \textbf{58.5} & \textbf{98.1} & \textbf{95.3} & \textbf{91.2} & \textbf{80.4} & 69.3 & \textbf{26.3} & 83.5 & \textbf{81.1} & \textbf{56.0} & 49.8 & \textbf{58.5} & 54.8 & \textbf{74.1} & \textbf{74.2} \\\bottomrule
\end{tabular}}
\vspace{-0.3cm}
\caption{\textbf{Linear probing.} We report on a variety of classification benchmarks with ViT-B/16 pre-trained on IT35M. CLIP achieves consistent performance gains compared to CLIP in a wide range of classification datasets.}
\label{tab:linearprobing}
\end{table*}
\subsection{Zero-Shot Transfer}
\label{sec:zeroshot}

\paragraph{Classification.}
We evaluate 27 classification benchmarks with zero-shot classification protocols following~\cite{clip}. As shown in \cref{tab:zeroshot}, \ourablmethod achieves an average top-1 accuracy of 32.7\%. Though performing 7.6\% worse than CLIP, \ourablmethod demonstrates descent zero-shot recognition capability without explicitly training with negative examples. \ourmethod{} achieves consistent performance gain across a wide range of datasets, with a 43.6\% average top-1 accuracy, which is \textbf{3.3}\% higher than CLIP. It indicates that the synergy between contrastive and non-contrastive objectives improves the model's zero-shot classification ability.

\paragraph{Out-of-domain classification.} For out-of-distribution classification, we use 5 datasets following~\cite{clip}: ImageNet Adversarial~\cite{imageneta}, ImageNet Rendition~\cite{imagenetr}, ImageNetV2~\cite{imagenetv2}, ImageNet Sketch~\cite{imagenetske}, and ObjectNet~\cite{objectnet} with 2 additional datasets: ImageNet-C~\cite{imagenetc} and Stylized ImageNet~\cite{imagenetsty}. We investigate if the non-contrastive objective inherits the robustness of the contrastive objective to natural distribution shift. As shown in \cref{tab:oodzeroshot}, \ourablmethod achieves an 18.0\% with CLIP an 23.1\% average top-1 accuracy, revealing a similar trend to in-domain datasets. \ourmethod{} further improves CLIP's accuracy by \textbf{3.6}\% and achieves an accuracy of 26.7\% across 7 datasets, indicating its effectiveness among out-of-domain datasets.

\paragraph{Multi-label classification.} We evaluate zero-shot multi-label classification on NUS-WIDE~\cite{nuswide} and OpenImages~\cite{openimages} following standard protocol~\cite{zsl}. Specifically, we use 81 unseen labels for NUS-WIDE and the most frequent 400 unseen test labels for OpenImages during evaluation, following~\cite{lesa}. As shown in \cref{tab:multilabel}, we observe that \ourablmethod shows the best mAP compared to CLIP. \ourablmethod is pre-trained without negative examples, resulting in the model's intrinsic strengths on recall over overlapped concepts, with 16.1 and 81.4 points of mAP on NUS-WIDE and OpenImages, respectively. Comparatively, \ourmethod{} slightly improves CLIP in this respect while lagging behind \ourablmethod. The experiments demonstrate that the non-contrastive objective fits well for those downstream tasks in demand of high recall.

\paragraph{Retrieval.} We evaluate on 2 retrieval benchmarks: Flickr30K~\cite{flickr30k} and MSCOCO~\cite{coco} under zero-shot protocol. We do not use prompt engineering and use the original caption. Compared to zero-shot classification, image-text retrieval requires the model's recognition capability at a fine-grained level. We empirically observe in \cref{tab:retrieval} that \ourablmethod performs drastically worse than CLIP, which is because text captions in retrieval are not mutually exclusive, which is different from label names in classification. Therefore, explicit negative examples during pre-training play an imperative role in zero-shot retrieval. See \cref{sec:failcase} for further discussions.  Beyond that, \ourmethod{} achieves a noticeable performance improvement over CLIP, with a gain of \textbf{3.7}\% R@1 on Flickr30K and \textbf{3.9}\% R@1 on MSCOCO, illustrating that the additional non-contrastive objective renders extra semantic signals that can be transferred well to fine-grained recognition.

\subsection{Representation Learning}
\label{sec:representation}

\begin{table}[tbp]
\centering
\setlength{\tabcolsep}{6mm}{
\begin{tabular}{cccc}
\multirow{2}{*}{Model} & \multicolumn{2}{c}{sm. sup.} & ft. \\
\cmidrule(lr){2-3}\cmidrule(lr){4-4}
& 1\% & 10\% & 100\% \\
\toprule
 CLIP & 57.8 & 72.4 & 82.4\\
\ourablmethod & 55.0 & 72.2 & 82.4 \\
\bf \ourmethod{} & \textbf{59.5} & \textbf{73.4} & \textbf{82.7} \\
\bottomrule
\end{tabular}}
\vspace{-0.2cm}
\caption{\textbf{Fine-tuning (ft.) and semi-supervised learning (sm. sup.).} Percentage shows sampling ratio for end-to-end fine-tuning. We report top-1 accuracy on the ImageNet-1K validation set. \ourablmethod performs comparatively with CLIP. \ourmethod{} induces consistent gains in all sampling ratios, especially for semi-supervised learning. }
\label{tab:finetune}
\end{table}

\begin{table}[tbp]
\centering
\setlength{\tabcolsep}{1.2mm}{
\begin{tabular}{cccccccc|c}
Model & \rotatebox{90}{\footnotesize STS12} &  \rotatebox{90}{\footnotesize STS13} & \rotatebox{90}{\footnotesize STS14} & \rotatebox{90}{\footnotesize STS15} & \rotatebox{90}{\footnotesize STS16} & \rotatebox{90}{\footnotesize STS-B} & \rotatebox{90}{\footnotesize SICK-R} & \rotatebox{90}{\footnotesize \bf Average} \\
\toprule
CLIP & \textbf{52.3} & 54.0 & 53.9 & 69.9 & 58.6 & \textbf{67.0} & 69.3 & 60.7 \\
\ourablmethod & 51.3 & \textbf{62.7} & \textbf{56.9} & \textbf{71.9} & \textbf{65.8} & 66.9 & 64.1 & \textbf{62.8} \\
\bf \ourmethod{} & 48.7 & 55.6 & 55.6 & 69.7 & 59.5 & 66.3 & \textbf{70.2} & 60.8 \\
\bottomrule
\end{tabular}}
\vspace{-0.2cm}
\caption{\textbf{Sentence embedding performance on semantic textual similarity.} We report on a variety of NLP understanding benchmarks. \ourmethod{} demonstrates the best capability for various STS tasks. \ourmethod{} has comparable performance with CLIP.}
\label{tab:sts}
\end{table}

We conduct several evaluation protocols to benchmark representation quality under different objectives for both visual (\cref{sec:visualrepresentation}) and textual (\cref{sec:textualrepresentation}) modality.

\subsubsection{Visual Representation Learning}
\label{sec:visualrepresentation}

\paragraph{Linear probing.} We evaluate the quality of visual representation via linear probing protocol, where a linear head is fine-tuned on top of the frozen backbones.  We follow the same setup as~\cite{mocov3} on the same 27 datasets as standard zero-shot classification. Specifically, we use SGD without momentum as the optimizer, no weight decay, and a total epoch of 100 for all evaluation datasets. We use \texttt{[CLS]} token for classification. Following~\cite{ibot}, we sweep a set of different learning rates by adding multiple classification heads over a shared frozen backbone, each with its own optimizer and scheduler. We report the best result across different heads.
As shown in \cref{tab:linearprobing}, we observe that \ourablmethod generally leads to comparable linear probing accuracy compared with CLIP, suggesting that non-contrastive objectives are able to derive semantically meaningful embedding spaces. \ourablmethod achieves 2.5\% higher performance compared to CLIP on ImageNet but relatively lower on average.
Beyond that, when combining two objectives together, \ourmethod{} achieves \textbf{2.7}\% higher performance on ImageNet and \textbf{1.5}\% higher performance on average across 27 different classification tasks compared to CLIP.

\paragraph{Fine-tuning \& semi-supervised learning.} We fine-tune the entire network under the full-data regime (100\%) and the semi-supervised learning protocol (1\% and 10\%) on ImageNet-1K~\cite{imagenet}. Results are shown in \cref{tab:finetune}. 
For CLIP and \ourmethod, we find that fine-tuning with the projection head trained under the contrastive objective yields consistent performance gain, especially under a low-data regime. Specifically, we take the text features of the label names as the classifier's initialized parameters. We opt for this setup by default. 
We highlight that while \ourablmethod performs worse than CLIP in terms of zero-shot transfer, the non-contrastive objective also serves as a strong baseline to learn representation considering the close gap between CLIP and \ourablmethod in terms of fine-tuning accuracy.
\ourmethod{} achieves an \textbf{1.7}\%, \textbf{1.0}\%, and \textbf{0.3}\% performance gain compared with CLIP under three data ratios of 1\%, 10\%, and 100\%, respectively.

\paragraph{Mask probing.} To evaluate how well the visual model is capable of deriving explicit scene layout and object boundaries, we conduct mask probing analysis following~\cite{dino}. For each attention head from the last layer, we extract the attention map with \texttt{[CLS]} token as the query. We then compute the Jaccard similarity $\mathcal{J}$ of each head's attention mask to the ground truth and retain the attention mask with the highest similarity. We conduct experiments on Pascal VOC 2012~\cite{pascalvoc} dataset.
With IT35M, \ourablmethod demonstrates better quality in the generated mask with 43.7 points of $\mathcal{J}$, while CLIP reaches 41.2 and \ourmethod{} reaches 41.9 points of $\mathcal{J}$. Hence, \ourablmethod learns better representation for object boundaries compared to CLIP, while \ourmethod{} strikes a balance between \ourablmethod and CLIP. Detailed results are delayed to \cref{sec:maskprobing}. As a related evaluation, we also study unsupervised segmentation with GroupViT~\cite{groupvit} in~\cref{sec:unsupseg}.

\subsubsection{Textual Representation Learning}
\label{sec:textualrepresentation}
We follow the setup as~\cite{simcse} and evaluate on 7 STS tasks: STS 2012–2016~\cite{sts12,sts13,sts14,sts15,sts16}, STS Benchmark~\cite{stsb}, and SICK-Relatedness~\cite{sickr}. We directly take \texttt{[EOS]} token without any projection as the input for evaluation, which consistently yields better performance compared to the projected feature for all models. We use cosine distance as the similarity metric. The main goal of sentence embeddings is to cluster semantically similar sentences, and hence, we take STS as one yardstick to benchmark textual representation learning.
Results are shown in \cref{tab:sts}. \ourablmethod shows better textual semantics compared to CLIP, with \textbf{2.1}\% higher in terms of the average score with IT35M. The conclusion can also be visually extrapolated from the t-SNE visualization of the word embedding, as detailed in~\cref{sec:tsne}.
We note that \ourmethod{} performs comparatively with CLIP given that STS is evaluated on \texttt{[EOS]} without projection, thus the backbone representation can be similar if dominant by the contrastive objective. 

\subsection{Properties}
\label{sec:properties}

\begin{table}[!tbp]
\centering
\setlength{\tabcolsep}{1.1mm}{
\begin{tabular}{lccccc}
\multirow{2}{*}{Data} & \multirow{2}{*}{Size} & \multirow{2}{*}{Ep.} & \multicolumn{3}{c}{ZS / LN Accuracy} \\
\cmidrule(lr){4-6}
& & & CLIP & \ourablmethod & \ourmethod{} \\
\toprule
\textcircled{\small{1}} & 12M & 25 & 36.8 / 68.5 & 37.5 / 71.0 & \textbf{42.4} / \textbf{72.2} \\
\textcircled{\small{4}} & \textcolor{gray!80}{14M} & \textcolor{gray!80}{32} & \textcolor{gray!80}{26.5 / 75.1} & \textcolor{gray!80}{\textbf{31.4} / 77.0} & \textcolor{gray!80}{27.1 / \textbf{77.2}} \\
\textcircled{\small{1}}\textcircled{\small{2}} & 22M  & 32 & 37.9 / 68.2 & 34.2 / \textbf{72.3} & \textbf{43.0} / 71.9 \\
\textcircled{\small{1}}\textcircled{\small{2}}\textcircled{\small{3}} & 35M  & 32 & 45.7 / 71.4 & 37.0 / 73.9 & \textbf{48.8} / \textbf{74.1} \\
\bottomrule
\end{tabular}}
\vspace{-0.2cm}
\caption{\textbf{Pre-training data scaling \& domain.} Ep. denotes training epochs. Data abbreviations are as follows. \textcircled{\small{1}}: CC12M. \textcircled{\small{2}}: COCO, VG, SBU, and CC3M. \textcircled{\small{3}}: YFCC14M. \textcircled{\small{4}}: IN21K. Note \textcircled{\small{1}}\textcircled{\small{2}}\textcircled{\small{3}} = IT35M.}
\label{tab:datascale}
\end{table}

\begin{figure}[!tbp]
\begin{minipage}[c]{.5\linewidth}
\centering
\includegraphics[height=1.0\linewidth]{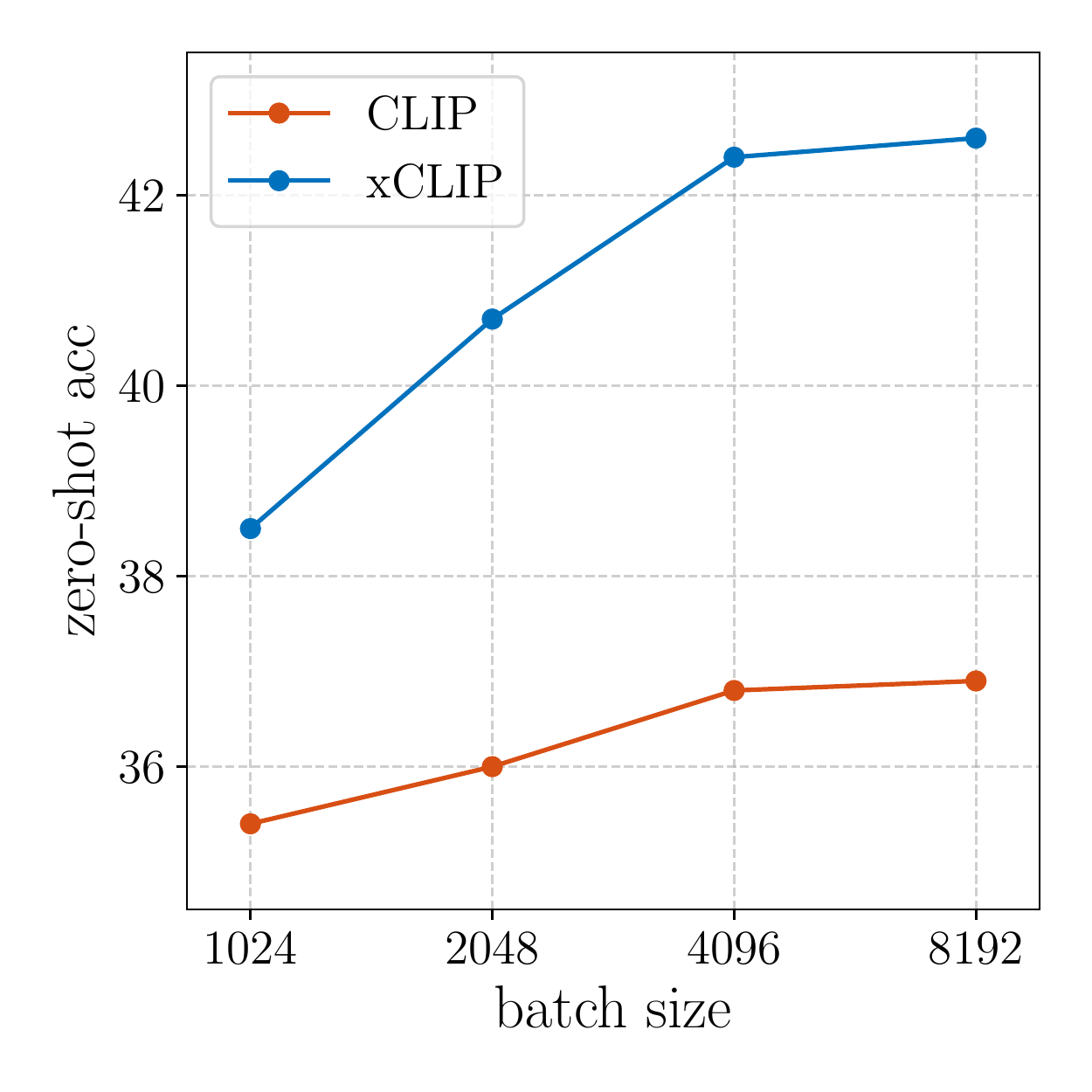}
\vspace{-0.8cm}
\captionsetup{width=.92\linewidth}
\caption{\textbf{Batch size scaling.} \ourmethod{} performs better than CLIP with a small batch size (\ie, 1024).}
\label{fig:batchsize}
\end{minipage}%
\begin{minipage}[c]{.5\linewidth}
\centering
\includegraphics[height=1.0\linewidth]{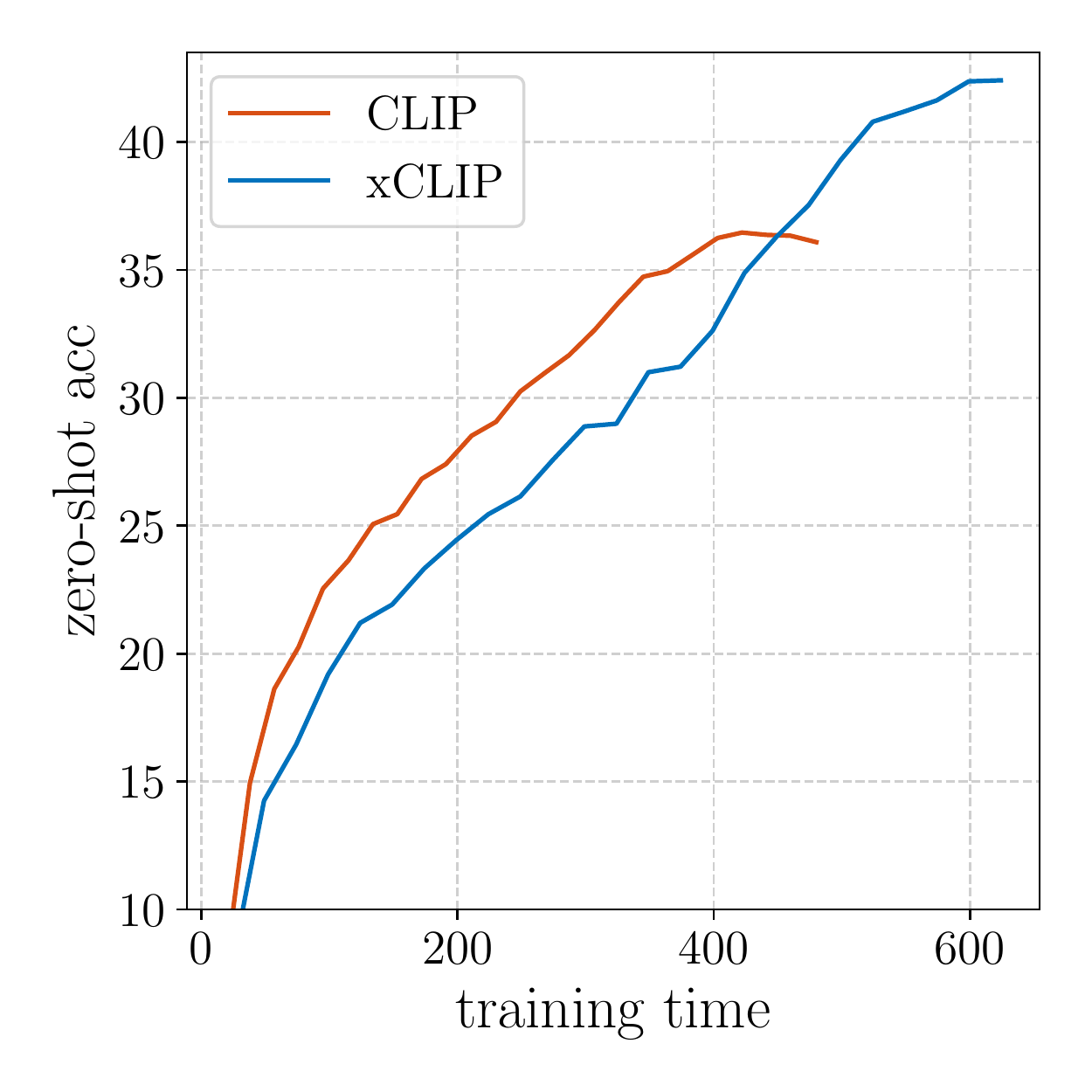}
\vspace{-0.8cm}
\captionsetup{width=.92\linewidth}
\caption{\textbf{Training time.} \ourmethod{} adds $\sim$1.3x training time cost and performs better with equal seen images.}
\label{fig:trainingtime}
\end{minipage}
\end{figure}

We analyze several crucial properties to demonstrate the generalizability and effectiveness of our method in practical usage.  Experiments are conducted with ViT-B/16 on CC12M for 25 epochs by default. 

\paragraph{Pre-training data scaling.} 
We study the model's performance with data scaling in \cref{tab:datascale}. 
\ourmethod{} exhibits desirable scaling behaviors similar to CLIP. The performance improves as data increases. However, we do not observe such a tendency in \ourablmethod in terms of zero-shot accuracy. For example, \ourablmethod achieves 37.5\% with \textcircled{\small{1}} but only 34.2\% with \textcircled{\small{1}}\textcircled{\small{2}}, the trend of which defies those of CLIP's (36.8\% \textit{vs.} 37.9\%) and \ourmethod's (42.4\% \textit{vs.} 43.0\%). 
On an important note, both \ourablmethod and \ourmethod{} showcase desirable scaling behaviors under the linear probing protocol. We hypothesize that the performance of zero-shot classification depends on pre-training data quality with the non-contrastive objective, while strong representation can always be ensured before the projection.

\paragraph{Pre-training with tagging data.}
We consider using \textit{tagging data}, \eg, ImageNet-21K~\cite{imagenet}, as pre-training data to validate method's generalizability. To do that, we take label names as annotation texts by concatenating them with a sampled prompt, similar to~\cite{basic,coca}. 
As shown in \cref{tab:datascale}, \ourablmethod shows superiority to CLIP when pre-trained with IN21K by performance gain of \textbf{4.9}\% and \textbf{1.9}\% in terms of zero-shot and linear probing accuracy, respectively. 
Compared to \ourablmethod, we instead notice a performance drop of 4.3\% for zero-shot accuracy when two objectives multi-tasked, indicating that the contrastive objective stymies the quality of projected probability under tagging data. 
Note, that an \textbf{77.2}\% linear probing accuracy achieved by \ourmethod, 2.1\% higher than CLIP, is decently strong compared to 79.0\% achieved by the state of the art in SSL, iBOT~\cite{ibot}, pre-trained using the same data with 80 epochs.

\paragraph{Batch size scaling.} We investigate the scaling behavior of CLIP and \ourmethod{} with different pre-training batch sizes. 
As shown in \cref{fig:batchsize}, \ourmethod{} performs stronger with smaller batch sizes than CLIP pre-trained even with larger batch sizes. For example, \ourmethod{} pre-trained with a batch size of 1024 achieves a zero-shot accuracy of 38.5\% (\textit{vs.} CLIP's 36.8\% with a batch size of 4096).
However, we still observe a substantial performance drop as batch size decreases, the behavior of which is different from self-supervised models pre-trained with the non-contrastive objective~\cite{swav,dino} that are insensitive to batch size. We hypothesize that a large batch size is necessary to approximate $\mathbb{E}(p)$ with $\frac{1}{B}\sum^B_{i=1}\vp_i$ in estimating $\mathcal{L}_\mathrm{HE}$, especially when two probabilities are derived from drastically different modalities.

\paragraph{Computation efficiency.} To demonstrate methods' computation efficiency, we show the FLOPs and GPU memory consumption of \ourmethod{} compared to CLIP. \ourmethod{} enforces bearable additional computation cost on top of CLIP with only 1.4\% extra FLOPs and 27\% extra GPU memory. As shown in \cref{fig:trainingtime}, \ourmethod{} adds $\sim$1.3x extra time cost but with a 6.0\% performance gain in terms of zero-shot accuracy. 

\subsection{Ablation Study}
\label{sec:ablation}

\begin{table*}[tbp]
\begin{subtable}[t]{0.33\linewidth}
\centering
\setlength{\tabcolsep}{3mm}{
\begin{tabular}[t]{llcc}
$\lambda_1$ & $\lambda_2$ &  ZS & LN \\
\toprule
\textcolor{gray!80}{0} & \textcolor{gray!80}{1} & \multicolumn{2}{c}{\textcolor{gray!80}{Nan}}\\
0.3 & 1 & 26.8 & 69.5 \\
0.5 & 1 & 23.7 & 67.2 \\
0.2 & 1.2 & 35.3 & 69.0 \\
\rowcolor{cyan!50} 0.5 & 1.5 & 37.5 & 71.0 \\
\bottomrule
\end{tabular}}
\captionsetup{width=.95\linewidth}
\caption{\textbf{Coefficient of entropy regularizer.} $\lambda_1 + 1 = \lambda_2$ generally leads to good performance.}
\label{tab:coefficient}
\end{subtable}
\vspace{1mm}
\begin{subtable}[t]{0.33\linewidth}
\centering
\setlength{\tabcolsep}{2mm}{
\begin{tabular}[t]{lccc}
dim &  ZS & LN & Mem \\
\toprule
8192 & 31.0 & 70.6 & 1.07$\times$ \\
16384 & 35.1 & 70.9 & 1.13$\times$ \\
\rowcolor{cyan!50} 32768 & 37.5 & 71.0 & 1.27$\times$ \\
65536 & 35.9 & 71.0 & 1.52$\times$ \\
\bottomrule
\end{tabular}}
\vspace{4mm}
\caption{\textbf{Projection dimension.} Mem (G) is compared to CLIP with a 512-dim embedding layer.}
\label{tab:projdim}
\end{subtable}
\begin{subtable}[t]{0.33\linewidth}
\centering
\setlength{\tabcolsep}{2.0mm}{
\begin{tabular}[t]{lcc}
arch &  ZS & LN \\
\toprule
\rowcolor{cyan!50} vanilla & 37.5 & 71.0 \\
w/ bottleneck layer & 31.2 & 69.0 \\
w/ l2-norm layer & 35.4 & 70.7 \\
w/o last BN layer & 34.3 & 69.9 \\
\bottomrule
\end{tabular}}
\vspace{4mm}
\captionsetup{width=.95\linewidth}
\caption{\textbf{Projection architecture} of non-contrastive branch. }
\label{tab:projarch}
\end{subtable}
\begin{subtable}[t]{0.33\linewidth}
\centering
\setlength{\tabcolsep}{3mm}{
\begin{tabular}[t]{lcc}
\#SL \ / \ \#TL &  ZS & LN \\
\toprule
 \rowcolor{cyan!50} 0 \ / \ 1 & 42.4 & 72.2 \\
 0 \ / \ 2 & 41.3 & 71.3 \\
 1 \ / \ 2 & 42.4 & 71.9 \\
 2 \ / \ 3 & 42.5 & 72.3 \\
\bottomrule
\end{tabular}}
\caption{\textbf{Shared projection layers.} \#SL denotes shared hidden layers. \#TL denotes total hidden layers.}
\label{tab:sharedlayers}
\end{subtable}
\begin{subtable}[t]{0.33\linewidth}
\centering
\setlength{\tabcolsep}{2.0mm}{
\begin{tabular}[t]{lcc}
$\lambda_\mathrm{CLIP}:\lambda_\mathrm{\ourablmethod}$ & ZS & LN \\
\toprule
1.0 & 40.2 & 66.5 \\
0.5 & 41.7 & 70.3 \\
\rowcolor{cyan!50} 0.2 & 42.4 & 72.2 \\
0.1 & 41.8 & 72.6 \\
\bottomrule
\end{tabular}}
\caption{\textbf{Loss ratio} between $\mathcal{L}_\mathrm{CLIP}$ and $\mathcal{L}_\mathrm{\ourablmethod}$. }
\label{tab:lossratio}
\end{subtable}
\begin{subtable}[t]{0.33\linewidth}
\centering
\setlength{\tabcolsep}{2.0mm}{
\begin{tabular}[t]{lcc}
technique & ZS & LN \\
\toprule
\rowcolor{cyan!50} vanilla & 42.4 & 72.2 \\
shared space (\cref{eq:onelatentxclip}) & 29.4 & 68.9 \\
debiased sampling & 42.2 & 72.4 \\
warm-up on $\mathcal{L}_\mathrm{\ourablmethod}$ & 41.7 & 71.9 \\
\bottomrule
\end{tabular}}
\captionsetup{width=.95\linewidth}
\caption{\textbf{Training objective and technique} for multi-tasking. }
\label{tab:taggingdataopt}
\end{subtable}
\vspace{-0.3cm}
\caption{\textbf{Ablation study} with ViT-B/16 on ImageNet-1K validation set. We report zero-shot (ZS) and linear probing (LN) accuracy (\%).}
\label{tab:ablation}
\end{table*}

We show in this section the crucial composing factors of \ourablmethod and \ourmethod. The ablations in the first column (\cref{tab:coefficient,tab:projdim,tab:projarch}) are conducted with $\mathcal{L}_\mathrm{\ourablmethod}$ only. The ablations in the second column (\cref{tab:lossratio,tab:sharedlayers,tab:taggingdataopt}) are conducted with full loss $\mathcal{L}_\mathrm{\ourmethod}$. Experiments are conducted with ViT-B/16 on CC12M for 25 epochs. The default settings are highlighted in \colorbox{cyan!50}{cyan}.

\paragraph{Entropy regularizer.} We study the effects of coefficient for entropy regularizers $\lambda_1$ and $\lambda_2$ in \cref{tab:coefficient}. The pre-training is unstable without regularizers or with only maximization of the entropy of the mean ($\mathcal{L}_\mathrm{HE}$). Specifically, the model will collapse to constant uniform distribution and outputs the same uniform probability distribution despite different inputs. The additional minimization of the mean of the entropy ($\mathcal{L}_\mathrm{EH}$) stabilizes training, but incurs dimensional collapse, eroding the performances (26.8\% and 23.7\% \textit{vs.} 37.5\% in terms of zero-shot accuracy). Simultaneously adjusting $\lambda_2$ with constraints $\lambda_1+1=\lambda_2$ and $\lambda_1=0.5$ mitigates the problem and yields the optimal performance, with a 37.5\% zero-shot and 71.0\% linear-probing accuracy. Details are shown in \cref{sec:entropyreg}. 

\paragraph{Projection head.} We study the design of two projection heads for $\mathcal{L}_\mathrm{CLIP}$ and $\mathcal{L}_\mathrm{\ourablmethod}$. As shown in \cref{tab:projdim}, large projection dimension matters for good performance. We opt for a projection dimension of 32768 balancing the performance and speed. As shown in \cref{tab:projarch}, the last BN layer is crucial and leads to more stable optimization in our experiments. We draw on the idea of prototypes~\cite{dino,paws} by introducing bottleneck and l2-norm layers, while they do not yield a performance gain. We further study whether two projections can share intermediate computation in \cref{tab:sharedlayers}. While sharing middle hidden layers with slightly deeper projection heads performs better, we opt for 2-layer MLP without shared layers for its simplicity.

\paragraph{Optimization.} We study the effect of hyper-parameters in optimization. 
In \cref{tab:lossratio}, we study the optimal loss ratio between $\mathcal{L}_\mathrm{CLIP}$ and $\mathcal{L}_\mathrm{\ourablmethod}$. 
We further study whether it's the best route for two objectives to be optimized in separate latent spaces via multi-tasking. Specifically, we consider a bespoke objective \cref{eq:onelatentxclip} taking account of probability estimation and negative samples simultaneously in a shared latent space. Detailed formulation are shown in \cref{sec:onelatent}. Optimizing in one latent space yields sub-optimal results with a nearly 13.0\% drop in zero-shot accuracy and 3.3\% in linear probing accuracy, suggesting two objectives are intrinsically contradictory. 
We consider debiased sampling, where each batch is sampled from a single data source, since their semantics may be closer and thus more suitable for optimization. This yields similar results.
We also add $\mathcal{L}_\mathrm{\ourablmethod}$ after warm-up epochs with the linear scheduler to its base scale, which erodes the performance.

\section{Conclusion}
\label{sec:conclusion}
Aligning images and texts is of overriding significance for vision-language understanding. To conquer the systematic insufficiency of the contrastive objective for acquiring semantics and tackling loose correlations between noisy image-text pairs, we explore the non-contrastive objective for language-image pre-training and unravel its properties. Empirical evidence reveals that the non-contrastive objective induces models to perform favorably in representation learning yet poorly in zero-shot transfer. Observing the distinct mechanisms of the two objectives, we further seek synergy between the two, and introduce a simple multi-tasking framework, \ourmethod, that enjoys the best of both worlds: \ourablmethod aids CLIP mining semantics while CLIP inherits intrinsic strengths for zero-shot recognition. The consistent performance gain of \ourmethod{} over CLIP on a wide variety of downstream tasks consolidates our findings. As potential future work, we may continue to scale up the data size (\eg, LAION400M~\cite{laion400m}) as well as the model size (\eg, ViT-L~\cite{vit}) to verify whether the scaling law applies and the performance improvement endures.

\clearpage
{\small
\bibliographystyle{ieee_fullname}
\bibliography{egbib}
}

\clearpage
\appendix

\section{Methodology Comparison}
\label{sec:methodcompare}

\begin{algorithm}[!bp]
\begin{minipage}{1.0\textwidth}
\BlankLine
\DontPrintSemicolon

\tcp{$\vf_I, \vf_T$: image \& text projection}

\tcp{$\vp_\cdot, \vq_\cdot$: target \& predicted probability}

\SetKwFunction{FMain}{$\mathcal{L}_\mathrm{unified}$}
\SetKwProg{Fn}{def}{:}{}
\Fn{\FMain{$\vf_I$, $\vf_T$, $\tau_s$, $\tau$, $\lambda_1$, $\lambda_2$}}{
    
    $\vp_I$ = $\mathrm{uniform}$(softmax($\vf_I$ / $\tau_s$, dim=1))
    
    $\vp_T$ = $\mathrm{uniform}$(softmax($\vf_T$ / $\tau_s$, dim=1))
    
    $\vq_I$ = softmax($\vf_I$ / $\tau$, dim=1)
    
    $\vq_T$ = softmax($\vf_T$ / $\tau$, dim=1)
    
    $\mathcal{L}_{\mathrm{CE}}$ = - ($\vp_I \cdot$ log($\vq_T$) + $\vp_T \cdot$ log($\vq_I$)).sum(dim=1)
    
    \hskip3em .mean(dim=0)

    $\mathcal{L}_{\mathrm{EH}}$ = - ($\vp_I \cdot$ log($\vp_I$) + $\vp_T \cdot$ log($\vp_T$)).sum(dim=1)
    
    \hskip3em .mean(dim=0)
    
    $\overline \vp_I$, $\overline \vp_T$ = $\vp_I$.mean(dim=0), $\vp_T$.mean(dim=0)
    
    $\mathcal{L}_{\mathrm{HE}}$ = - ($\overline \vp_I \cdot$ log($\overline \vp_I$) + $\overline \vp_T \cdot$ log($\overline \vp_T$)).sum(dim=1)

    $\mathcal{L}$ = $\mathcal{L}_{\mathrm{CE}}$ + $\lambda_1 \cdot \mathcal{L}_{\mathrm{EH}}$ - $\lambda_2 \cdot \mathcal{L}_{\mathrm{HE}}$
    
    \KwRet $\mathcal{L}$ / 2
    \BlankLine
}
\caption{PyTorch-like pseudo-code of the unified \\ objective for non-contrastive pre-training.}
\label{alg:method}
\end{minipage}
\end{algorithm}

\begin{table}[!tbp]
\begin{center}
\setlength{\tabcolsep}{1.0mm}{
\begin{tabular}{lcccccc}
\multirow{2}{*}{Method} & \multicolumn{2}{c}{Sharpness} & \multicolumn{2}{c}{Smoothness} & \multirow{2}{*}{ZS} & \multirow{2}{*}{LN} \\
\cmidrule(lr){2-3}\cmidrule(lr){4-5}
& $\tau/\tau_s$ & $\lambda_1$ & $\mathrm{uniform}(\cdot)$ & $\lambda_2$ & & \\
\toprule
\textcolor{gray!80}{CE} & \textcolor{gray!80}{1} & \textcolor{gray!80}{0} & \textcolor{gray!80}{-} & \textcolor{gray!80}{0} & \multicolumn{2}{c}{\textcolor{gray!80}{Nan}} \\
\cmidrule(lr){2-7}
SwAV~\cite{swav} & 1 / 0.25 & 0 & Sinkhorn & 1$^*$ & 27.9 & 70.3 \\
SCSF~\cite{selfclassifier} & 1 / 0.5 & 0 & Batch-Softmax & 0 & 26.5 & 70.0 \\
DINO~\cite{dino} & 1 / 0.7 & 0 & Centering$^\dag$ & 0 & 22.9 & 69.1 \\
\cmidrule(lr){2-7}
\multirow{2}{*}{MSN~\cite{msn}} & 1 / 0.8 & 0 & - & 1 & \multicolumn{2}{c}{Nan} \\
 & 1 / 0.7 & 0 & - & 1 & 37.4 & 70.8 \\
\cmidrule(lr){2-7}
& 1 / 0.8 & 0.2 & - & 1.2 & 37.4 & 70.0 \\
\rowcolor{cyan!50}\multirow{-2}{*}{\ourablmethod}\cellcolor{white} & 1 & 0.5 & - & 1.5 & 37.5 & 71.0 \\
\bottomrule
\end{tabular}}
\end{center}
\vspace{-0.5cm}
\caption{\textbf{Comparison over different non-contrastive objectives.} CE denotes vanilla cross-entropy. ZS and LN denote top-1 zero-shot and linear probing accuracy. $^*$ The original SwAV~\cite{swav} implementation sets $\lambda_2$ as 0 while it is necessary under our setup to set $\lambda_2$ as 1 to avoid collapsing solutions. $^\dag$ Centering is performed in practice before (instead of after) softmax.}
\label{tab:methodcompare}
\end{table}

In this section, we compare different non-contrastive objectives derived from cross-entropy loss with different techniques to avoid model collapse, \emph{i.e.}, to ensure both the \textit{sharpness} and \textit{smoothness} of the representation. We first showcase a general formulation to achieve this end in \cref{alg:method}, and substantiate different self-supervised techniques proposed in the vision field to our setting. We tuned hyper-parameters for each item to suit best for language-image pre-training. We follow the proposition of non-collapsing representations~\cite{msn,paws}, where 
\begin{equation}
\label{eq:noncollpasing}
||\nabla {\overline H}_\theta(\vp_i) || + ||\nabla H_\theta(\overline \vp) || > 0
\end{equation}
$\forall i \in [B]$ if the network parameters $\theta$ lead to collapsing representation, \ie, $\vp_j = \vp_k$, $\forall j, k \in [B]$. $\vp_i \in \R^K$ is a K-dimensional probability distribution for i$^\mathrm{th}$ instance. In both cases where $\vp_i=\frac{1}{K}\bm{1}_K$ or $\vp_i\neq\frac{1}{K}\bm{1}_K$, the above equation holds such that the unified objectives is immune to collapsing representations.

Experiments are conducted with ViT-B/16 on CC12M for 25 epochs, the results of which are listed in \ref{tab:methodcompare}. For Centering, we use additional running variance and scale parameters to uniform the target probability, which yields more stable training compared to the original~\cite{dino} with a running mean only. For objectives with an explicit function $\mathrm{uniform}(\cdot)$, the asymmetry of the forward pass of the target and predicted probability incurs an inconsistency between training and evaluation, which leads to poor zero-shot performance. While the descent linear probing accuracy of these objectives suggests capabilities to learn representation, they lose a unique advantage to perform zero-shot recognition. 
MSN~\cite{msn} uses the mean entropy maximization regularizer to ease the usage of explicit uniform function, which is essentially the $\mathrm{HE}$ term.
Comparatively, the joint optimization of $\mathrm{EH}$ and $\mathrm{HE}$ does not rely on different temperatures for target and predicted probability, creating perfect symmetry between the two branches, and is observed to also improve the training stability (Nan in row \red{5} \textit{vs.} 37.0\% in row \red{7}).

\section{Additional Ablations}
\label{sec:addablation}

We provide additional ablation studies in this section. Experiments are conducted with ViT-B/16 on CC12M for 25 epochs by default.

\begin{figure*}[!t]
\centering
\begin{subfigure}[b]{0.22\textwidth}
\centering
\includegraphics[width=\textwidth]{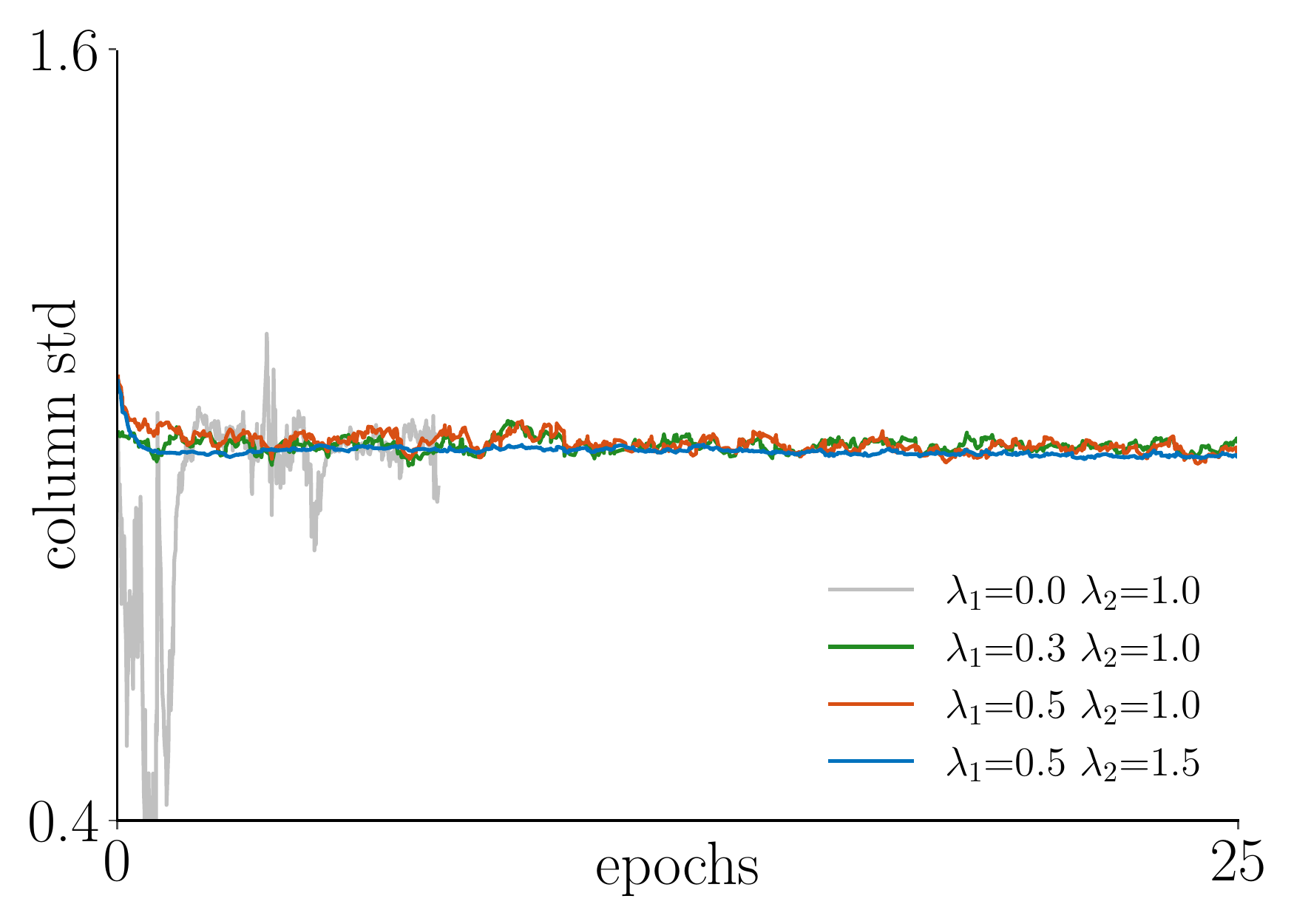}
\caption{\textbf{Column std.} Std converges to 1 if training is stable due to Batch Normalization. }
\label{fig:columnstd}
\end{subfigure}
\hspace{1.7cm}
\begin{subfigure}[b]{0.22\textwidth}
\centering
\includegraphics[width=\textwidth]{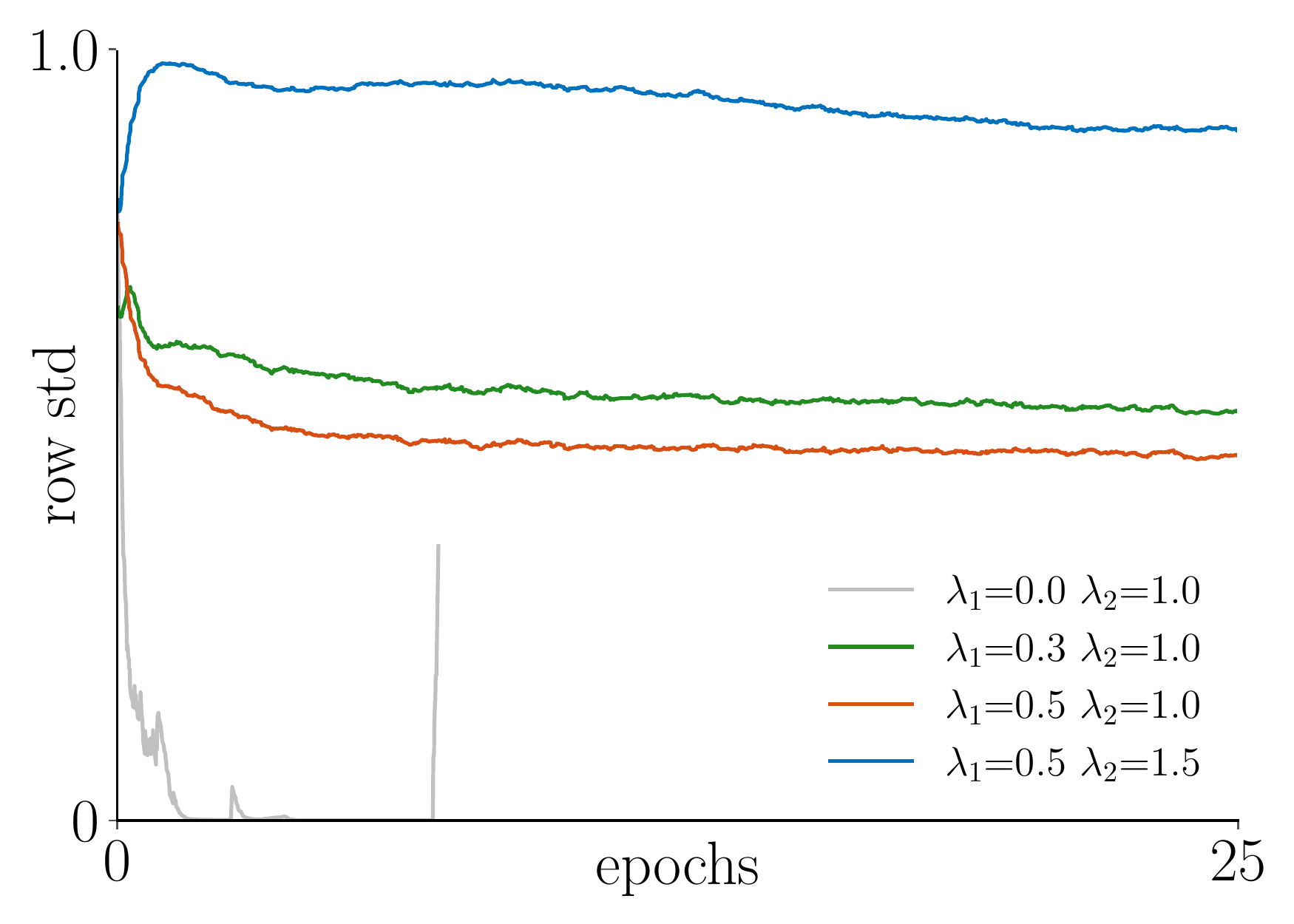}
\caption{\textbf{Row std.} An unevenly large $\lambda_1$ yields a small row std. $\lambda_2$ should be increased accordingly.}
\label{fig:row_std}
\end{subfigure}
\hspace{1.7cm}
\begin{subfigure}[b]{0.22\textwidth}
\centering
\includegraphics[width=\textwidth]{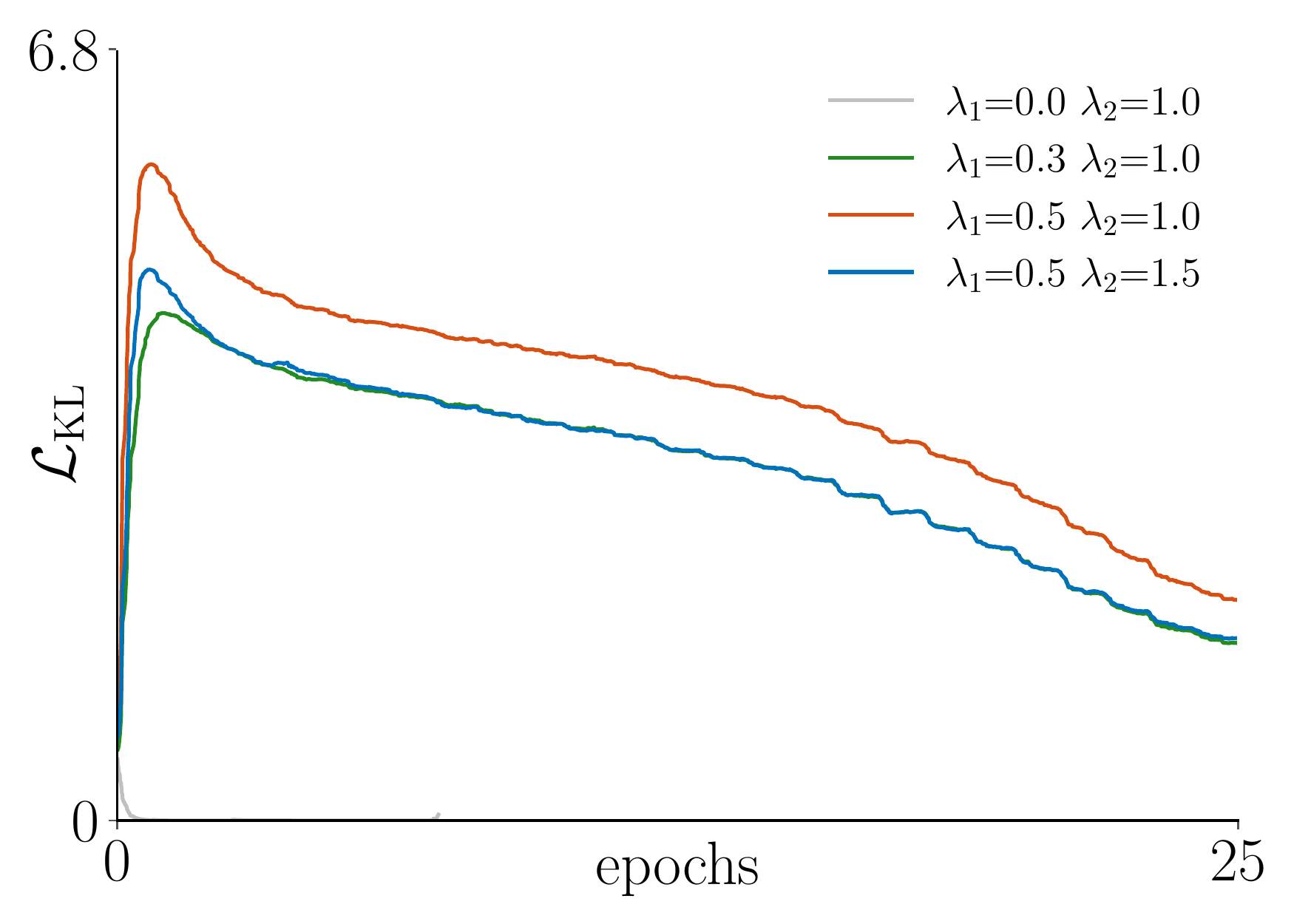}
\caption{\textbf{$\mathcal{L}_\mathrm{KL}$.} The discrepancy between the two modalities increase at first if the training is stable.}
\label{fig:kl}
\end{subfigure} \\
\vspace{0.1cm}
\begin{subfigure}[b]{0.22\textwidth}
\centering
\includegraphics[width=\textwidth]{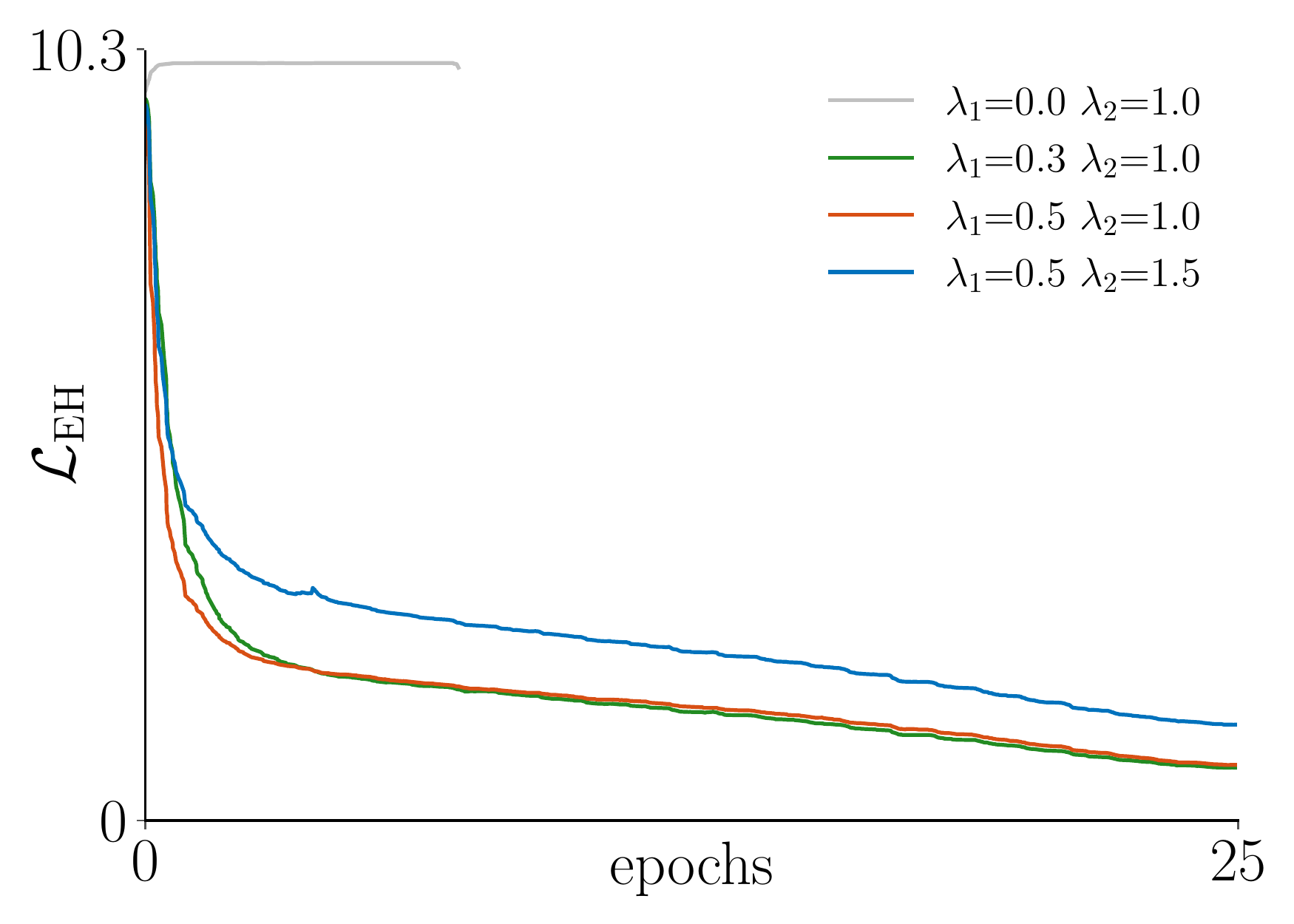}
\caption{\textbf{$\mathcal{L}_\mathrm{EH}$.} Entropy is minimized to be close to 0. Imbalanced or larger $\lambda_1$ yields smaller $\mathcal{L}_\mathrm{EH}$.}
\label{fig:eh}
\end{subfigure} 
\hspace{1.7cm}
\begin{subfigure}[b]{0.22\textwidth}
\centering
\includegraphics[width=\textwidth]{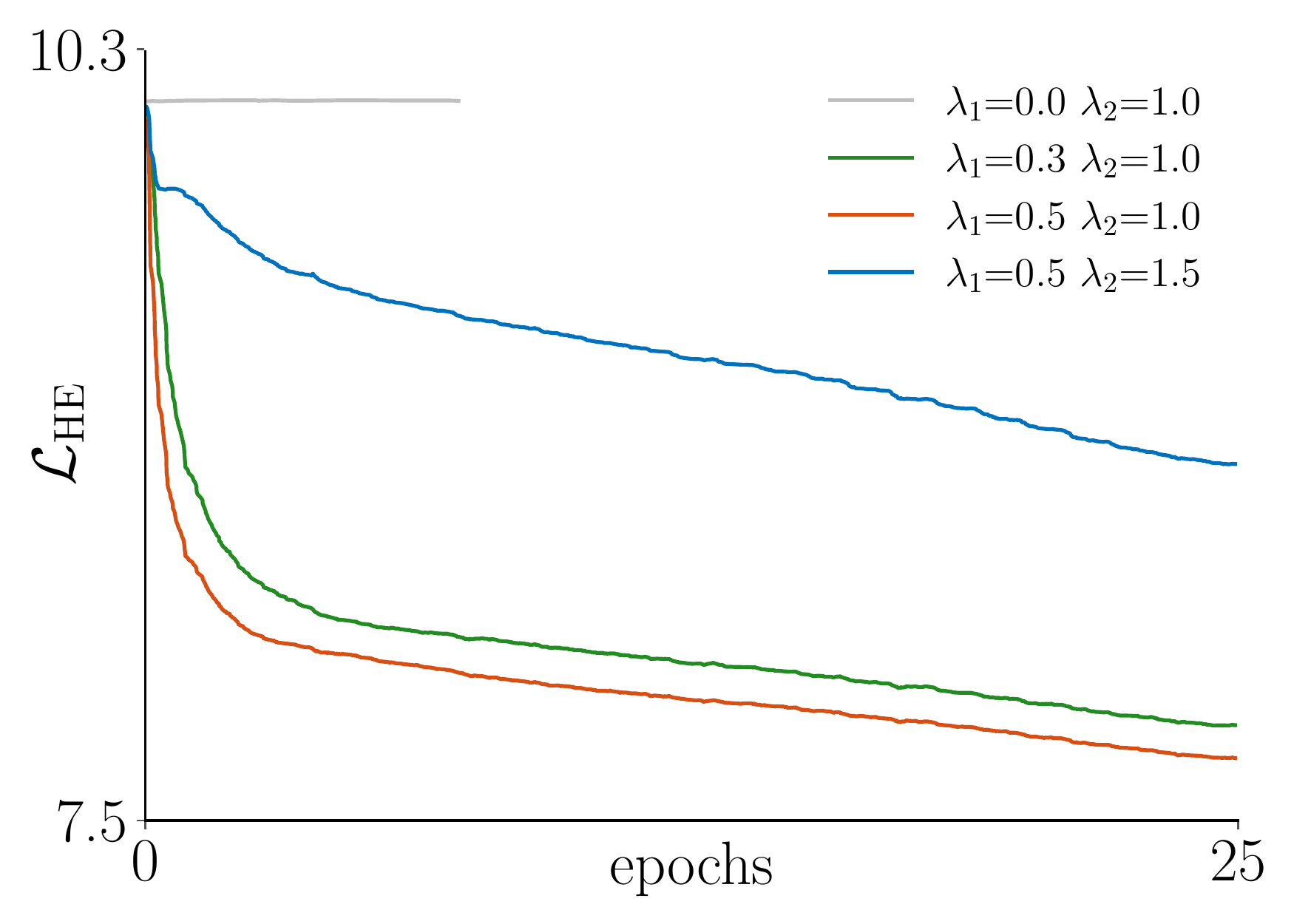}
\caption{\textbf{$\mathcal{L}_\mathrm{HE}$.} Mean entropy is maximized to ensure a uniform distribution. }
\label{fig:he}
\end{subfigure}
\hspace{1.7cm}
\begin{subfigure}[b]{0.22\textwidth}
\centering
\includegraphics[width=\textwidth]{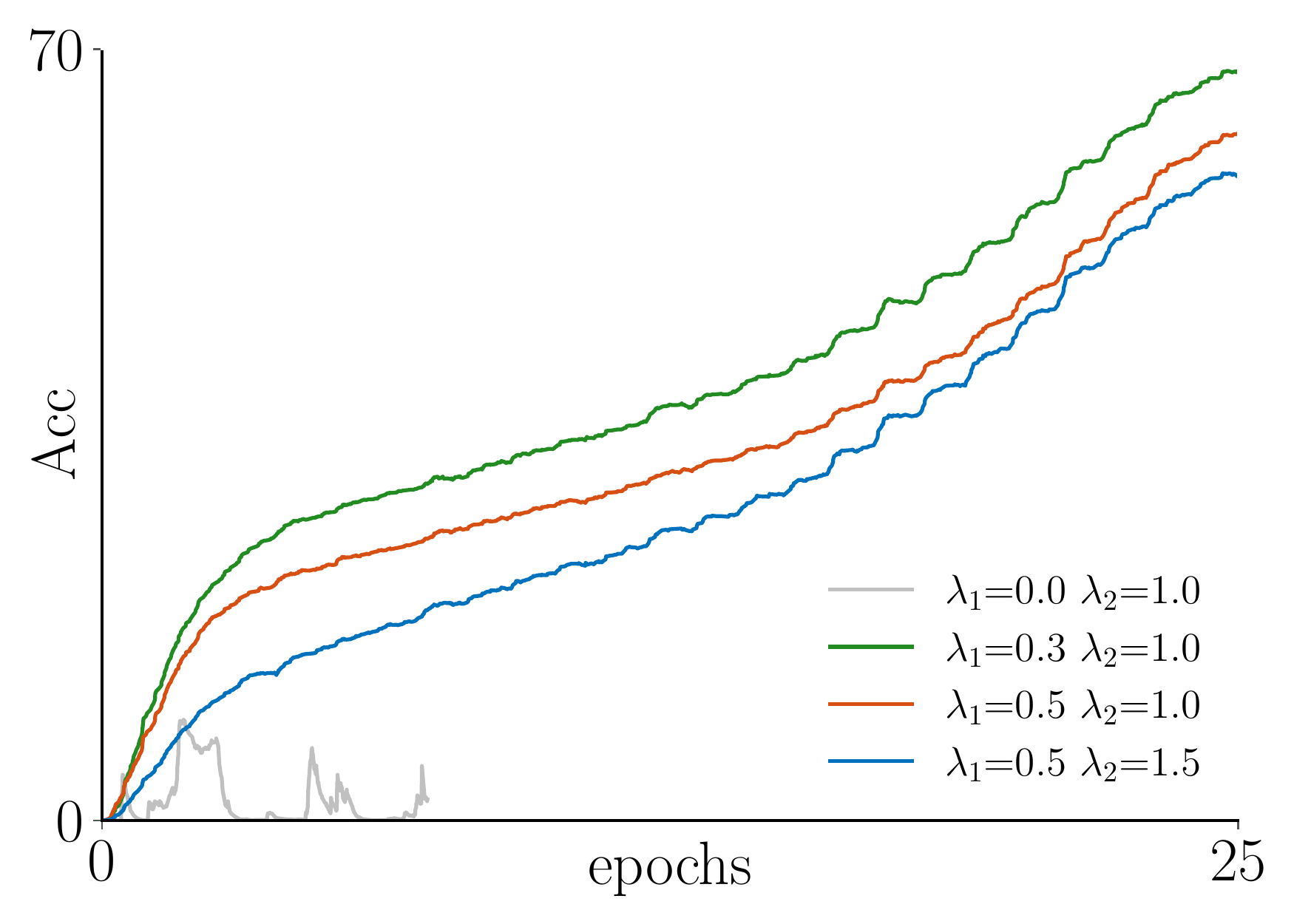}
\caption{\textbf{Accuracy.} Accuracy measures how well the one-hot prediction from two modalities is matched. }
\label{fig:acc}
\end{subfigure}
\vspace{-0.2cm}
\caption{\textbf{Entropy regularizer.} We showcase the training statistics with different $\lambda_1$ and $\lambda_2$. Setting $\lambda_1>0$ is the key to avoiding collapsing solutions. $\lambda_1=0.5$ and $\lambda_2=1.5$ leads to stable training and optimal downstream performance. }
\label{fig:entropy}
\vspace{-0.3cm}
\end{figure*}

\subsection{Entropy Regularizer}
\label{sec:entropyreg}
In this section, we testify claims made on \cref{sec:ablation} using pre-training statistics as clues. We monitor the loss scale of $\mathcal{L}_\mathrm{KL}$(=$\mathcal{L}_\mathrm{CE}$-$\mathcal{L}_\mathrm{EH}$), $\mathcal{L}_\mathrm{EH}$, and $\mathcal{L}_\mathrm{HE}$, as well as the standard deviation (std) of the row and the column of probability matrix $\mP^\mathrm{T} = [\vp_1^\mathrm{T}, \vp_2^\mathrm{T}, ..., \vp_B^\mathrm{T}] \in \R^{B \times K}$. Note that $\mathcal{L}_\mathrm{CE} + \lambda_1 \cdot \mathcal{L}_\mathrm{EH} - \lambda_2 \cdot \mathcal{L}_\mathrm{HE} = \mathcal{L}_\mathrm{KL}  + (1+\lambda_1) \cdot \mathcal{L}_\mathrm{EH} - \lambda_2 \cdot \mathcal{L}_\mathrm{HE}$. With the results and statistics given in \cref{tab:coefficient} and \cref{fig:entropy}, we present the following discussions:
\begin{enumerate}[label=\textbf{\arabic*)},wide=10pt]

\item $\bf{\lambda_1=0, \lambda_2=1}$ (\textcolor{gray}{gray}). When using only $\mathcal{L}_\mathrm{HE}$, the model fails to converge. The row std decrease to 0 and column std oscillates drastically. $\mathcal{L}_\mathrm{EH}$ and $\mathcal{L}_\mathrm{HE}$ do not decrease, while $\mathcal{L}_\mathrm{KL}$ remains 0. This indicates that the distributions are collapsing to a \textit{constant uniform distribution}.

\item $\bf{\lambda_1>0, \lambda_2=1}$ (\textcolor{ggreen}{green} \& \textcolor{gorange}{orange}). If $\lambda_1>0$ throughout training, the column std will be less oscillating at the beginning. Both the row std and $\mathcal{L}_\mathrm{KL}$ will increase rapidly. This mitigates the instability at the start. However, the row std will be undesirably low after some iterations, this is due to \textit{dimensional collapse}, where several dimensions will not be favored by any instance. 

\item $\bf{\lambda_1=\lambda_2-1>0}$ (\textcolor{gblue}{blue}). This couples $\mathcal{L}_\mathrm{EH}$ and $\mathcal{L}_\mathrm{HE}$ into one term $\lambda_2 \cdot (\mathcal{L}_\mathrm{EH} - \mathcal{L}_\mathrm{HE})$, optimizing less on $\mathcal{L}_\mathrm{KL}$ while more on non-collapsing condition, which stabilizes the training at the beginning when two modalities can be drastically different. $\mathcal{L}_\mathrm{KL}$ arises more rapidly at the beginning. Increasing $\lambda_2$ eases the dimensional collapse at the latter part of the training.
\end{enumerate}

\begin{table}[!tbp]
\begin{center}
\setlength{\tabcolsep}{2.3mm}{
\begin{tabular}{lcccccc}
Method & $\lambda_1$ & $\lambda_2$ & Acc$_\mathrm{C}$ & Acc$_\mathrm{nC}$ & ZS & LN \\
\toprule
\textcolor{gray!80}{CLIP} & \textcolor{gray!80}{-} & \textcolor{gray!80}{-} & 88.3 & - & \textcolor{gray!80}{36.8} & \textcolor{gray!80}{68.5} \\
\textcolor{gray!80}{nCLIP} & \textcolor{gray!80}{-} & \textcolor{gray!80}{-} & - & 59.0 & \textcolor{gray!80}{37.5} & \textcolor{gray!80}{71.0} \\
\cmidrule(lr){2-7}
\multirow{3}{*}{\oursecablmethod} & 0 & 0  & 81.6 & 0.1 & 38.9 & 69.9 \\
& 0.5 & 1.5 & 83.2  & 0.1 & 40.2 & 70.3 \\
& 3.0 & 4.0 & 66.3 & 29.3 & 29.4 & 68.9 \\
\cmidrule(lr){2-7}
\rowcolor{cyan!50}\ourmethod{}\cellcolor{white} & 0.5 & 1.5 & 89.2 & 60.2 & 42.4 & 72.2 \\
\bottomrule
\end{tabular}}
\end{center}
\vspace{-0.5cm}
\caption{\textbf{Meeting strategy of CLIP and \ourablmethod.} They are met in one latent space (\oursecablmethod) via a hybrid objective or separate latent spaces (\ourmethod) via multi-tasking of vanilla objectives. Acc$_\mathrm{C}$ denotes the accuracy of CLIP to discriminate positive samples across the batch. Acc$_\mathrm{nC}$ denotes the accuracy of nCLIP to predict the same one-hot assignment across the channel.}
\label{tab:onelatent}
\end{table}

\subsection{Meeting in One Latent Space}
\label{sec:onelatent}

We experiment with another idea where contrastive and non-contrastive objectives meet in one shared latent space instead of two separate latent spaces in a multi-task manner. Specifically, we introduce negative samples to the non-contrastive objective by replacing \cref{eq:celoss} to the matrix multiplication form with $\mathrm{InfoNCE}$. Let $\mP = [\vp_1, \vp_2, ..., \vp_B]$ and $\mQ = [\vq_1, \vq_2, ..., \vq_B]$. We consider the following $\mathrm{CE}$ term:
\begin{align}
\label{eq:onelatentce}
\mathcal{L}_\mathrm{\tilde{CE}} &= \mathrm{InfoNCE}((\mP^\mathrm{T} \mathrm{log}(\mQ) + \mathrm{log}(\mP)^\mathrm{T} \mQ)/ \sigma) \nonumber \\[5pt]
& = - \frac{1}{B} \sum_{i\in B} \mathrm{log}\frac{\mathrm{exp}((\vp_i^\mathrm{T}\mathrm{log}(\vq_i)+\mathrm{log}(\vp_i)^\mathrm{T}\vq_i)/ \sigma)}{\sum_j\mathrm{exp}((\vp_i^\mathrm{T}\mathrm{log}(\vq_j)+\mathrm{log}(\vp_i)^\mathrm{T}\vq_j)/ \sigma)} \nonumber \\[5pt]
& = - \frac{1}{B} \sum_{i\in B} ((\vp_i^\mathrm{T}\mathrm{log}(\vq_i)+\mathrm{log}(\vp_i)^\mathrm{T}\vq_i)/ \sigma - \Phi(Z_{i,\cdot})) \nonumber \\[-1pt]
& = \mathcal{L}_\mathrm{CE} / \sigma - \frac{1}{B} \sum_{i\in B} \Phi(Z_{i,\cdot}).
\end{align}
$\Phi$ denotes log-sum-exponential $\mathrm{log}\sum_j\mathrm{exp}(x_j)$ and is an approximation of $\mathrm{max(\cdot)}$ function with $\sigma \to 0$. Compared to \cref{eq:celoss} that only computes losses over the diagonal terms of the distance matrix, we now also optimize over the off-diagonal terms. This ensures unmatched negative image-text pairs are distracted under the distance metric of cross-entropy, while matched positive pairs are attracted. The final objective is 
\begin{equation}
\label{eq:onelatentxclip}
\mathcal{L}_\mathrm{\oursecablmethod} = \mathcal{L}_\mathrm{\tilde{CE}} + \lambda_1 \cdot \mathcal{L}_\mathrm{EH} - \lambda_2 \cdot \mathcal{L}_\mathrm{HE}
\end{equation}
We would like to see whether the two objectives conflict and can be optimized in on latent space. As shown in \cref{tab:onelatent}, when $\lambda_1=\lambda_2=0$, $\mathcal{L}_\mathrm{\oursecablmethod}$ reduces to $\mathcal{L}_\mathrm{CLIP}$ with sole difference on distance metric and incurs a 2.1\% and 1.4\% performance gain on ZS and LN, respectively.
When $\lambda_1=\lambda_2-1>0$, $\mathcal{L}_\mathrm{\oursecablmethod}$ behaves more similar to $\mathcal{L}_\mathrm{\ourablmethod}$ with scaling and an additional negative term. Setting $\lambda_1=0.5, \ \lambda_1=1.5$ is insufficient to avoids model collapse as implied by Acc$_\mathrm{nC}$. While increasing the weights of regularizers and setting $\lambda_1=3, \ \lambda_1=4$ yields seemingly balanced Acc$_\mathrm{C}$ and Acc$_\mathrm{nC}$, we observe only a 29.4\% zero-shot accuracy with the pre-trained model, validating the contradiction of two objectives in one shared latent space. While a bespoke design may exist, we opt for simple multi-tasking of CLIP and \ourablmethod in separate spaces.

\section{Additional Implementation}
\label{sec:addimplement}

\paragraph{Zero-shot classification. } We follow the same setup as~\cite{clip}, with prompt engineering for each of the 27 evaluation datasets, including Food-101~\cite{food101}, CIFAR-10~\cite{cifar}, CIFAR-100~\cite{cifar}, Birdsnap~\cite{birdsnap}, SUN397~\cite{sun397}, Stanford Cars~\cite{cars}, FGVC Aircraft~\cite{aircraft}, Pascal VOC 2007 Classification~\cite{pascalvoc}, Describable Textures~\cite{DTD}, Oxford-IIIT Pets~\cite{pets}, Caltech-101~\cite{caltech101}, Oxford Flowers 102~\cite{flowers}, MNIST~\cite{mnist}, Facial Emotion Recognition 2013~\cite{fer2013}, STL-10~\cite{stl10}, EuroSAT~\cite{eurosat}, RESISC45~\cite{resisc45}, GTSRB~\cite{gtsrb}, KITTI~\cite{kitti}, Country211~\cite{clip}, PatchCamelyon~\cite{pcam}, UCF101~\cite{ucf101}, Kinetics700~\cite{k700}, CLEVR Counts~\cite{clevercounts}, Hateful Memes~\cite{hatefulmemes}, Rendered SST2~\cite{clip}, and ImageNet~\cite{imagenet}. The final text embedding is ensembled by averaging all text embeddings with different prompts. 

\paragraph{Fine-tuning \& semi-supervised learning. } We use a training recipe from~\cite{beit}, with a layer-wise learning rate decay rate of 0.65, a weight decay of 0.05, a drop path rate of 0.1, a total epoch of 100, and DeepSpeed. We use \texttt{[CLS]} token for classification. We disable relative positional embedding and layer scaling. We sweep over four learning rates \{$3e^{-3}$, $4e^{-3}$, $5e^{-3}$, $6e^{-3}$\} for all models. For semi-supervised learning with partial data, we find that keeping CLIP's projection head yields better performance, especially with 1\% of data. Specifically, the \texttt{[CLS]} token is further forwarded to CLIP's image projection head and is classified via a cosine classifier with the temperature learned during pre-training. we initialize the weight of the classifier as the text embedding of ImageNet labels. We do not observe evident gain when fine-tuning on full data.

\section{Additional Results}
\label{sec:addresult}

\begin{table}[!tbp]
\begin{center}
\setlength{\tabcolsep}{3.4mm}{
\begin{tabular}{lccc}
Model & Data & Supervision & $\mathcal{J}$ \\
\toprule
\textcolor{gray!80}{random} & \textcolor{gray!80}{-} & \textcolor{gray!80}{-} & 25.7 \\
DeiT~\cite{deit} & ImageNet & class & 30.4 \\
MSN~\cite{msn} & ImageNet & self & 38.6 \\
TWIST~\cite{twist} & ImageNet & self & 44.1 \\
iBOT~\cite{ibot} & ImageNet & self & 44.1 \\
DINO~\cite{dino} & ImageNet & self & 44.7 \\
MoCoV3~\cite{mocov3} & ImageNet & self & \textbf{45.9} \\
\midrule
CLIP & IT35M & text & 41.2 \\
\ourablmethod & IT35M & text & 43.7 \\
\textbf{\ourmethod} & IT35M & text & 41.9 \\
\bottomrule
\end{tabular}}
\end{center}
\vspace{-0.5cm}
\caption{\textbf{Masking probing.} $\mathcal{J}$ denotes Jaccard similarity between predictions and the ground-truth. Models with ViT-B/16 are listed. }
\label{tab:maskprobing}
\end{table}

\subsection{Mask Probing}
\label{sec:maskprobing}

The results are shown in \cref{tab:maskprobing}. For the top panel, we showcase the performance of a range of self-supervised models pre-trained with ImageNet-1K. MoCoV3~\cite{mocov3} achieves a $\mathcal{J}$ of 45.9 points while the supervised baseline~\cite{deit} lags behind with only a $\mathcal{J}$ of 30.4 points. For the bottom panel, we showcase text-supervised models pre-trained with different objectives. We note that text supervision can also derive explicit boundary scene layout, with all models achieving decent results. Among them, \ourablmethod performs the best with 43.9 points which is on par with advanced self-supervised models. 

\begin{table}[H]
\begin{center}
\setlength{\tabcolsep}{2.4mm}{
\begin{tabular}{lccc}
Model & Data & mAcc & mIoU \\
\toprule
CLIP & IT35M & 42.9 & 24.8 \\
\textbf{\ourmethod} & IT35M & \textbf{53.1} & \textbf{38.4} \\
\bottomrule
\end{tabular}}
\end{center}
\vspace{-0.5cm}
\caption{\textbf{Unsupervised segmentation with GroupViT.} Models based on ViT-S are evaluated. }
\label{tab:unsup}
\end{table}

\subsection{Unsupervised Segmentation with GroupViT}
\label{sec:unsupseg}

We consider small-size GroupViT~\cite{groupvit} under the setting of unsupervised semantic segmentation with PASCAL VOC 2012~\cite{pascalvoc} dataset. We strictly follow the original setup to train CLIP and a similar recipe to train \ourablmethod, with only one difference initializing learning rate as $5e^{-4}$. We do not use the multi-label loss as in~\cite{groupvit} for simplicity. The results are shown in \cref{tab:unsup}. \ourmethod{} achieves 38.4 points of mIoU and 53.1 points of mAcc, which is sufficiently better than CLIP baseline, indicating the non-contrastive objective help to learn better object boundaries and scene layouts.

\section{Visualization}
\label{sec:visualization}

\begin{figure*}[!tbp]
\centering
\includegraphics[height=0.5\linewidth,width=1\linewidth]{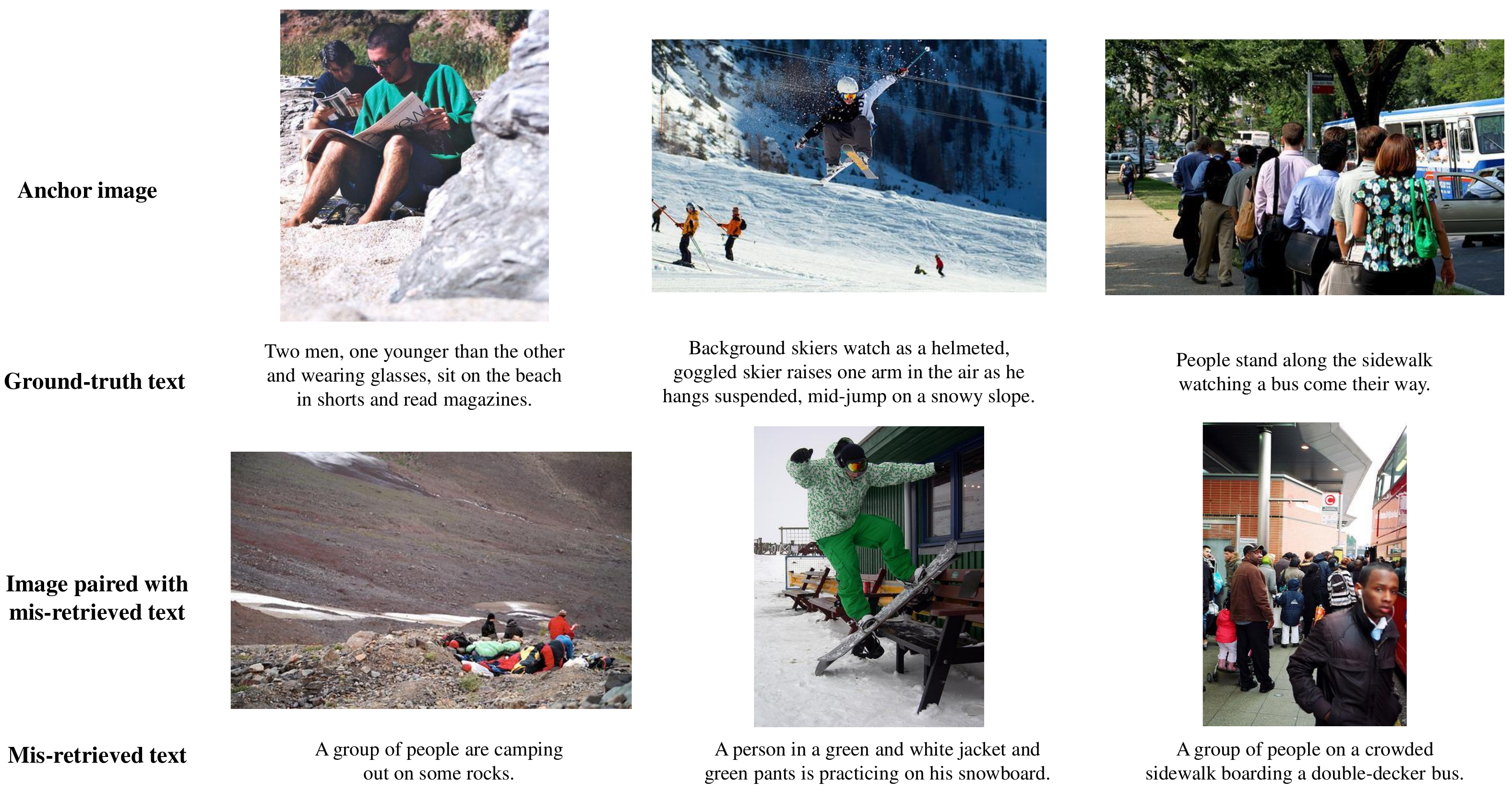}
\caption{\textbf{Failure case analysis} of \ourablmethod on zero-shot image-to-text retrieval. }
\label{fig:failcase}
\end{figure*}

\subsection{Failure Cases of \ourablmethod in Retrieval}
\label{sec:failcase}

We demonstrate several failure cases of \ourablmethod (that is correctly retrieved by CLIP) in zero-shot image-to-text retrieval. As shown in \cref{fig:failcase}, \ourablmethod tends to mis-retrieve texts containing similar visual elements (\eg, \textit{rocks} in the first column). Their paired images usually contain similar objects, conceptions, and scene layouts. Empirically, \ourablmethod tends to overlook fine-grained features, \ie, color, attribute, or number but remains generally good sense in predicting high-level semantics. In these senses, the non-contrastive objective serves better as an appending regularizer instead of one single term when downstream tasks solicits direct fine-grained projections as in zero-shot retrieval.

\section{Hyper-Parameters}
\label{sec:hyperparam}

\begin{table}[H]
\centering
\setlength{\tabcolsep}{4mm}{
\begin{tabular}[t]{lc}
Hyper-parameter & Value \\
\toprule
batch size & 4096 \\
training epochs & 32 \\
learning rate & $1e^{-3}$ \\
learning rate end & $2e^{-6}$ \\
learning rate scheduler & cosine decay \\
weight decay & 0.2  \\
warm-up epochs & 3 \\
optimizer & AdamW \\
Adam $\beta_1$ & 0.9\\
Adam $\beta_2$ &0.98 \\
Adam $\eps$ & $1e^{-6}$ \\
$\lambda_1$ & 0.5 \\
$\lambda_2$ & 1.5 \\
head arch & 4096 - 32768 \\
\bottomrule
\end{tabular}}
\captionsetup{width=.95\linewidth}
\caption{\textbf{Pre-training hyper-parameters.} }
\label{tab:hyperparams}
\end{table}

\subsection{Class Representation}
\label{sec:tsne}
\begin{figure*}[!t]
\centering
\includegraphics[width=1\linewidth]{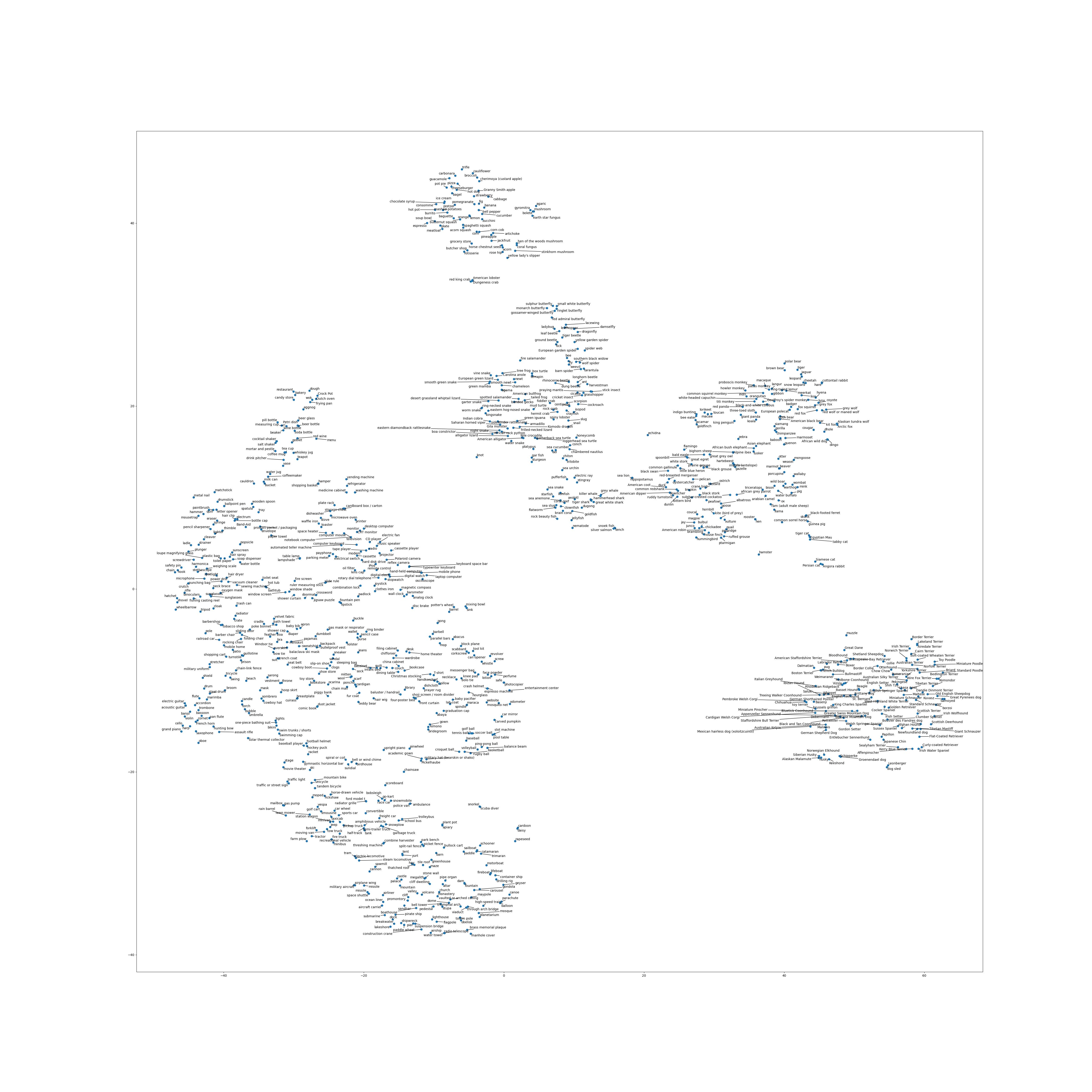}
\caption{\textbf{t-SNE visualization} on image embeddings of CLIP.}
\label{fig:tsneclipimage}
\end{figure*}

\begin{figure*}[!t]
\centering
\includegraphics[width=1\linewidth]{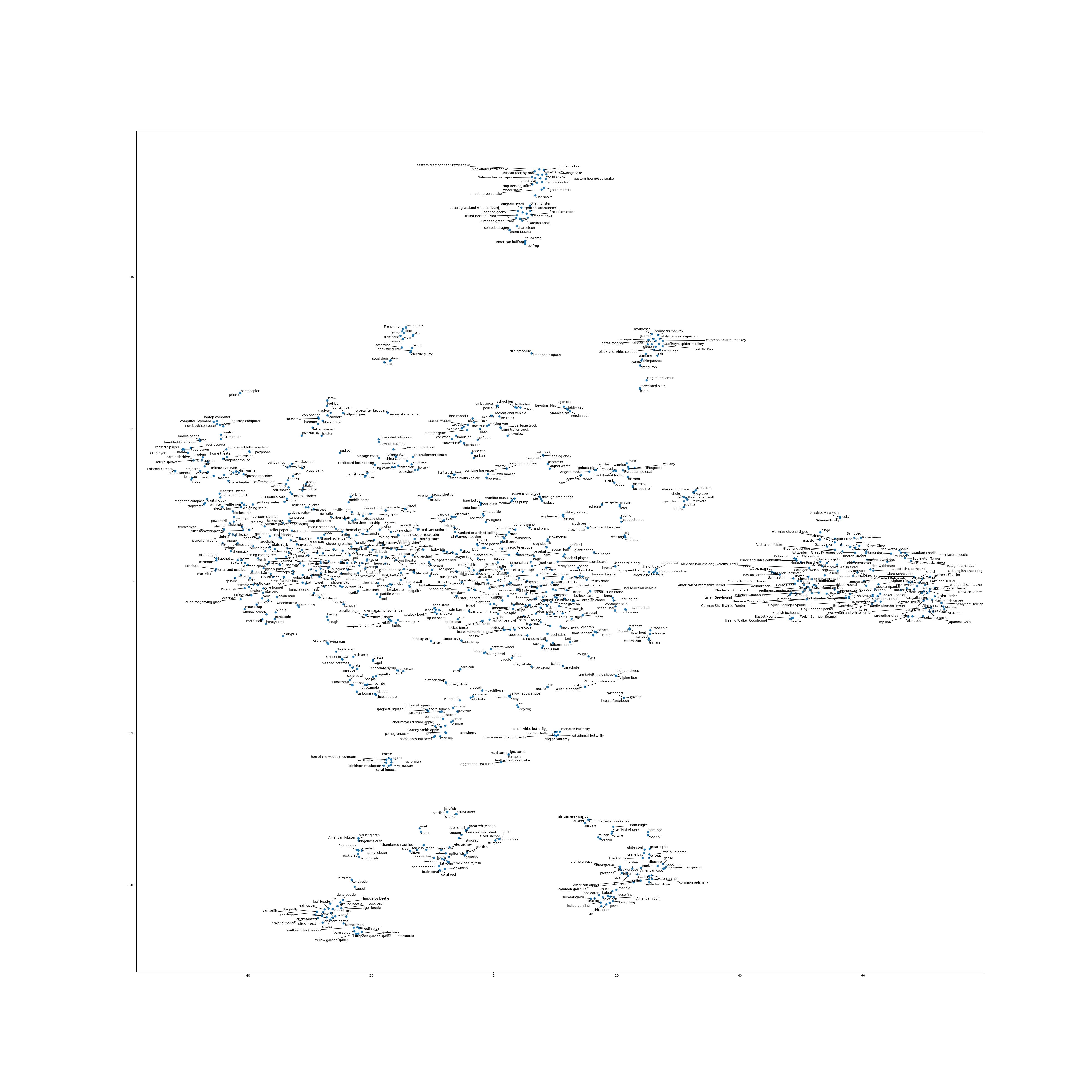}
\caption{\textbf{t-SNE visualization} on image embeddings of \ourablmethod.}
\label{fig:tsnenclipimage}
\end{figure*}

\begin{figure*}[!t]
\centering
\includegraphics[width=1\linewidth]{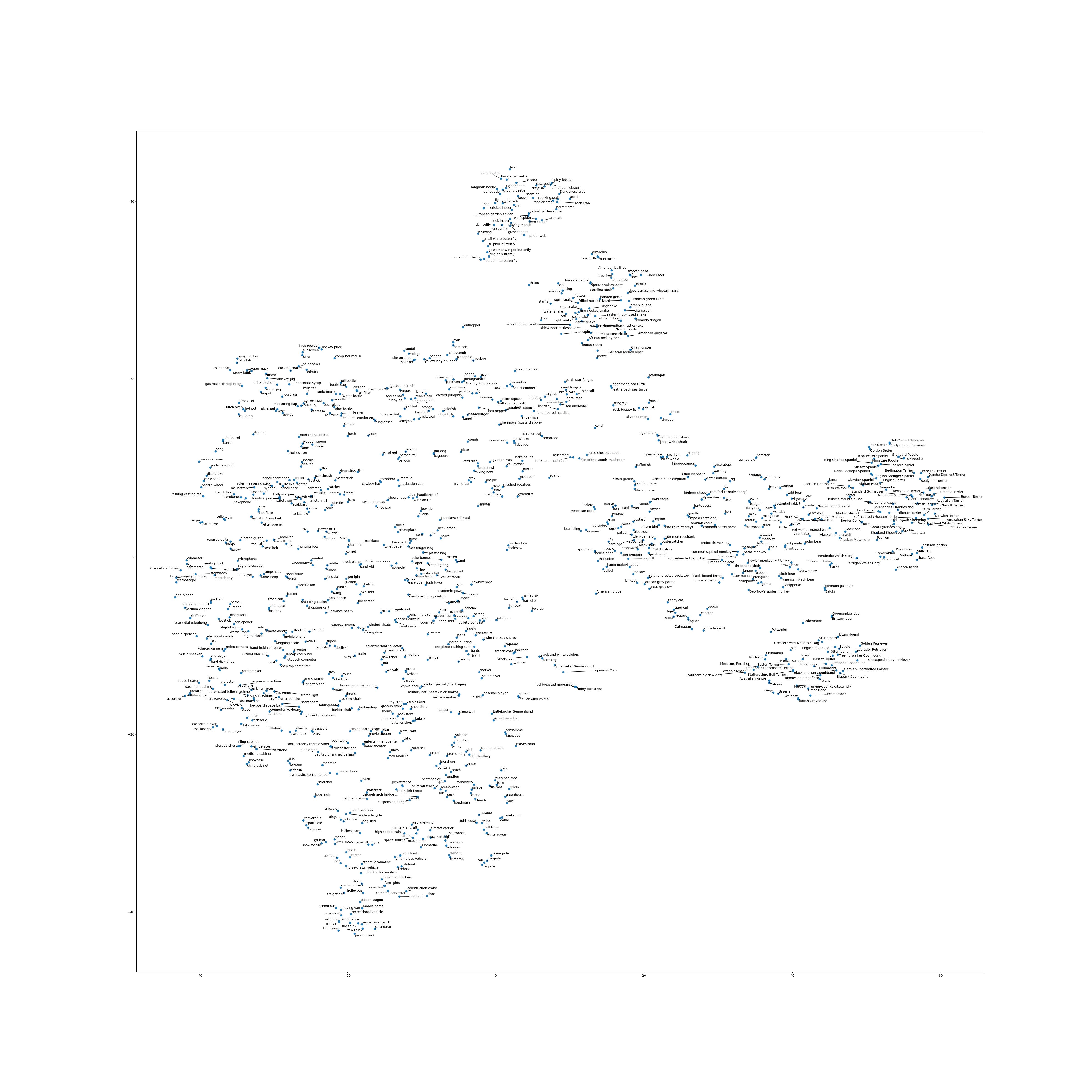}
\caption{\textbf{t-SNE visualization} on text embeddings of CLIP.}
\label{fig:tsnecliptext}
\end{figure*}

\begin{figure*}[!t]
\centering
\includegraphics[width=1\linewidth]{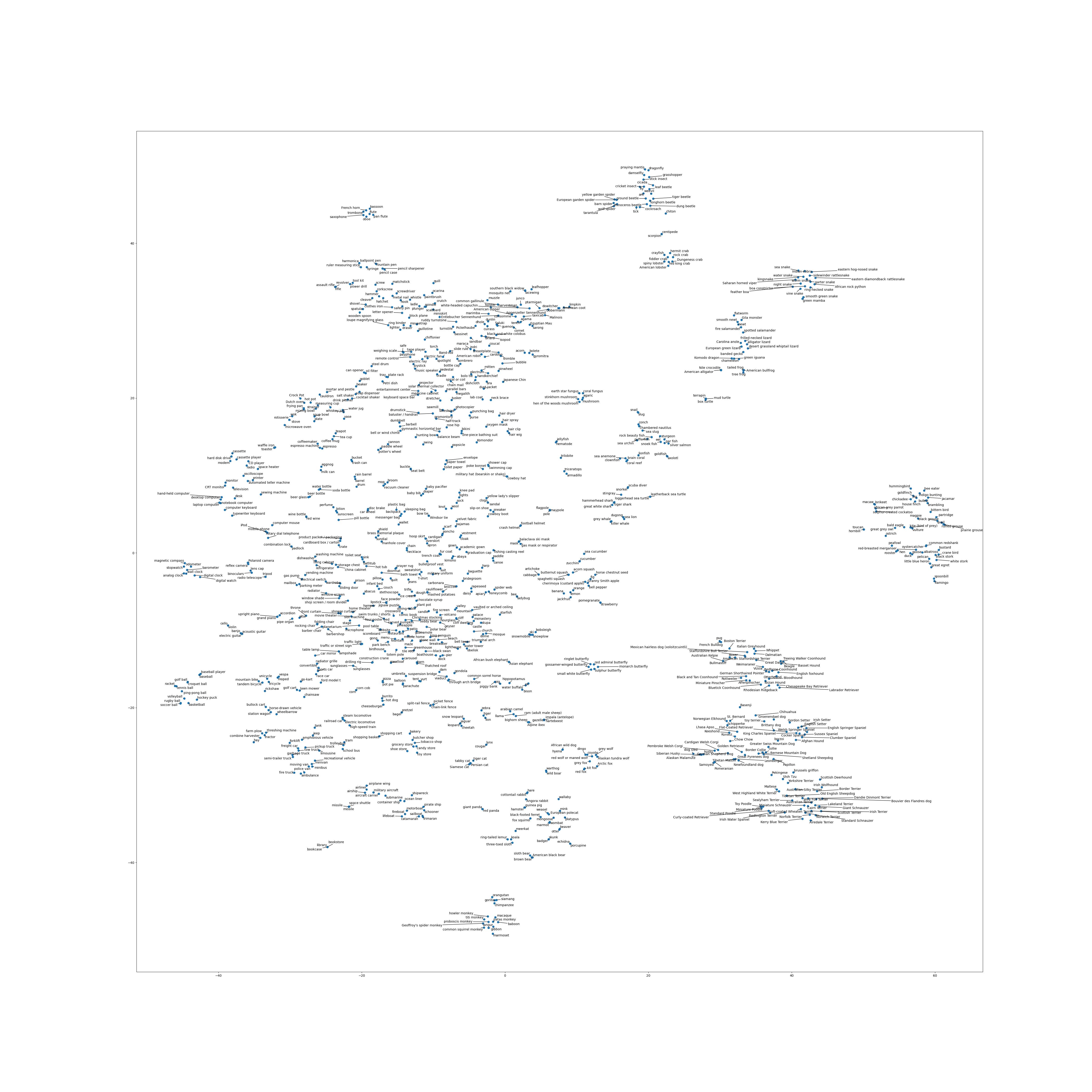}
\caption{\textbf{t-SNE visualization} on text embeddings of \ourablmethod.}
\label{fig:tsnencliptext}
\end{figure*}

To validate the non-contrastive term helps learning better semantics-meaningful representation, we visualize the t-SNE~\cite{tsne} of ImageNet-1K classes over validation set in \cref{fig:tsneclipimage,fig:tsnecliptext,fig:tsnenclipimage,fig:tsnencliptext}. Specifically, we showcase both the text's embedding and an average of images' embedding for each of the 1000 classes. We use cosine distance and CE as pre-computed metrics for contrastive and non-contrastive objectives, respectively. We run t-SNE with a perplexity of 20 and a learning rate of 200 for 5000 iterations. The non-contrastive objective comparatively leads to more semantic-meaningful clusters. For example, animals with different species are better separated and visually-similar objects are better attracted.

\end{document}